\documentclass[fontsize=11pt,parskip=full,dvipsnames]{article}
\usepackage{amsmath,amssymb,amsfonts,amsthm,mathabx}
\usepackage[dvips]{graphics}
\usepackage{enumerate}
\usepackage{mathrsfs}
\usepackage{subfigure}
\usepackage{color}
\usepackage{setspace}  %
\usepackage{hyperref}
\usepackage{float}
\usepackage{microtype} %
\usepackage{xfrac} %

\usepackage[font=small,skip=5pt]{caption} %

\usepackage[T1]{fontenc}

\usepackage{authblk} %
\usepackage{bbm} %

\usepackage{enumerate} %

\usepackage{amsmath,amssymb,amsfonts,amsthm,mathabx}
\usepackage{enumerate}
\usepackage{mathrsfs}
\usepackage{bbm}
\usepackage{subfigure}
\usepackage{hyperref}
\usepackage{url}
\usepackage{amsfonts}
\usepackage{xcolor}
\usepackage{graphicx}

\newcommand{\LR}{\text{LR}}

\usepackage{makecell}
\usepackage{xurl}
\usepackage{float}
\usepackage{multirow}

\usepackage[backend=biber, style=apa,url=false,isbn=false,uniquelist=false]{biblatex}
\AtEveryBibitem{\clearfield{month}}
\AtEveryBibitem{\clearfield{day}}
\AtEveryBibitem{\clearfield{language}}

\DeclareSymbolFont{AMSb}{U}{msb}{m}{n}

\usepackage{geometry}
\usepackage{url}

\usepackage{amsfonts}
\usepackage{xcolor}
\usepackage{graphicx}

\title{\Large{\textbf{LGDE: Local Graph-based Dictionary Expansion}}
}

\author[1,*]{Juni Schindler}
\author[2]{Sneha Jha}
\author[3,4]{Xixuan Zhang}
\author[3,4]{Kilian Bühling}
\author[3,4,5]{Annett Heft}
\author[1,**]{Mauricio Barahona}
\affil[1]{\normalsize{Department of Mathematics, Imperial College London,  United Kingdom}}
\affil[1]{\normalsize{Department of Surgery \& Cancer, Imperial College London, United Kingdom}}
\affil[3]{\normalsize{Weizenbaum Institute Berlin, Germany}}
\affil[4]{\normalsize{Institute for Media and Communication Studies, Free University of Berlin, Germany}}
\affil[5]{\normalsize{Institute for Research on Far Right Extremism, University of Tübingen,  Germany}}

\affil[*]{Corresponding author: \small\texttt{juni.schindler19{\scriptsize @}imperial.ac.uk}}

\affil[**]{Corresponding author: \small\texttt{m.barahona{\scriptsize @}imperial.ac.uk}}

\date{}

\begin{document}\maketitle

\begin{abstract}

  We present Local Graph-based Dictionary Expansion (LGDE), a method for data-driven discovery of the semantic neighbourhood of words using tools from manifold learning and network science. At the heart of LGDE lies the creation of a word similarity graph from the geometry of word embeddings followed by local community detection based on graph diffusion. The diffusion in the local graph manifold allows the exploration of the complex nonlinear geometry of word embeddings to capture word similarities based on paths of semantic association, over and above direct pairwise similarities. Exploiting such semantic neighbourhoods enables the expansion of dictionaries of pre-selected keywords, an important step for tasks in information retrieval, such as database queries and online data collection.  We validate LGDE on two user-generated English-language corpora and show that LGDE enriches the list of keywords with improved performance relative to methods based on direct word similarities or co-occurrences. We further demonstrate our method through a real-world use case from communication science, where LGDE is evaluated quantitatively on the expansion of a conspiracy-related dictionary from online data collected and analysed by domain experts. Our empirical results and expert user assessment indicate that LGDE expands the seed dictionary with more useful keywords due to the manifold-learning-based similarity network.
\end{abstract}

\noindent{\slshape\bfseries Keywords.}  dictionary expansion; word embeddings; cosine similarity; manifold learning; graph diffusion; local community detection
\section{Introduction}

\textit{Dictionary expansion} aims to expand a set of pre-selected keywords by adding related terms that can enhance original queries in information retrieval tasks. Designing a dictionary without in-depth knowledge of the vocabulary in the domain of interest is prone to inaccurate, non-specific or incomplete results. Therefore, expert-generated seed dictionaries are typically expanded with domain-specific keywords for diverse applications, such as patent searches~\autocite{leeTechnologyOpportunityIdentification2014}, queries of bibliometric databases~\autocite{yinCooccurrenceBasedApproach2020} and online forums~\autocite{gharibshahIKEAUnsupervisedDomainspecific2022}, query expansion for more effective web searches~\autocite{royUsingWordEmbeddings2016, kuziQueryExpansionUsing2016}, or collecting topic-specific content from social media platforms~\autocite{brunsCorona5GBoth2020,zengConceptualizingDarkPlatforms2021,klingerFringesMainstreamPolitics2022,atteveldtCAVAOpenSource2022a}. Dictionary expansion is particularly relevant, and challenging, in domains with evolving semantics where word choice and language style are highly specialised and diverge from general usage or are constantly in flux to ensure exclusive, community-internal communication, to adjust to ongoing events or the emergence of topics and cultural changes, or to avoid legal prosecution~\autocite{heftChallengesApproachesData2023}. Approaches based on large static corpora such as WordNet~\autocite{fellbaum1998wordnet} are unable to capture semantic change, which usually requires large diachronic corpora collected over long periods of time and are not ideal for highly dynamic linguistic environments such as social media.

When retrieving information around a certain topic, the challenge becomes to find a `good' dictionary that leads to a corpus containing most documents associated with the topic (high recall) and with few irrelevant documents (high precision). New approaches for data-driven dictionary expansion have leveraged word embedding models to find semantically similar words to pre-selected keywords~\autocite{royUsingWordEmbeddings2016, amslerUsingLexicalSemanticConcepts2020,gharibshahIKEAUnsupervisedDomainspecific2022,atteveldtCAVAOpenSource2022a,stollDevelopingIncivilityDictionary2023}. %

In this work, we build on the idea of data-driven dictionaries, but rather than focusing only on words that are most directly similar to pre-selected keywords, we introduce \textit{Local Graph-based Dictionary Expansion} (LGDE), a method that incorporates tools from manifold learning and network science to capture the complex geometry of the space of word embeddings. Previous studies~\autocite{jakubowskiTopologyWordEmbeddings2020} have suggested that the space of word embeddings is close to a `pinched' manifold within a high-dimensional ambient space. 
Our methodology contains two main ingredients: the representation of the complex geometry of word vector embeddings through a similarity graph, and the detection of local graph communities around seed words using random walks.  
Specifically, we start with domain-specific word vector embeddings and employ the \textit{Continuous $k$-Nearest Neighbors} (CkNN) geometric graph construction~\autocite{berryConsistentManifoldRepresentation2019} to create a semantic similarity network of words.
We then expand the pre-selected set of keywords by adding the words that belong in their corresponding \textit{semantic communities} as determined by \textit{Severability}~\autocite{yuSeverabilityMesoscaleComponents2020}, a fast local community detection method
that exploits random walks to explore the local nonlinear geometry of the word graph to find local communities around seed keywords, 
and thus naturally includes weighted graph paths corresponding to multi-step word associations.
Importantly, the CkNN graph provides a consistent description in the sense that its unnormalised graph Laplacian, which determines the properties of graph diffusions, converges to the Laplace-Beltrami operator in the limit of large data. This 
intrinsic consistency linking geometry and diffusion further motivates %
our use of severability, a diffusion-based method for local community detection, and we show it provides a quantitative advantage over other local community detection algorithms.

To evaluate our method, we consider the task of expanding a dictionary of pre-selected keywords from a human-coded hate speech English-language dataset from the social media platforms Reddit~(\texttt{https://reddit.com})
and Gab~(\texttt{https://gab.com}) 
\autocite{qianBenchmarkDatasetLearning2019}. When compared to baseline approaches based on direct word similarities or co-occurrences, LGDE leads to improved dictionary expansion, as measured by a higher $F_1$ score of discovered words significantly more likely to appear in hate speech-related communication. In addition, to showcase the general-purpose applicability of LGDE, we use a text classification setting on the 20 Newsgroups dataset~\autocite{mitchellTwentyNewsgroups1997}, also showing an improvement over baseline methods. Finally, we apply our method to a real-world 
scenario, where LGDE is used to aid subject matter experts in analysing conspiracy-related posts from the message forum 4chan (\texttt{https://4chan.org}) %
on the `Great Replacement'
and `New World Order'
conspiracy theories. In this case, LGDE shows a quantitative advantage in discovering additional relevant words that would be missed without the %
graph-based perspective, as assessed by domain expert-based (blind) annotation.

\subsection{Problem definition}
Let us consider a pre-selected list of $n$ keywords $W_0=\{w_1,w_2,...,w_n\}$, denoted the \textit{seed dictionary}, which are known to be relevant to a certain topic. 
These seed terms are typically derived from expert knowledge, literature research or existing keyword lists~\autocite{gharibshahIKEAUnsupervisedDomainspecific2022,heftChallengesApproachesData2023}. 
Let us further assume that we have access to a domain-specific corpus of $l$ documents $D=\{d_1, d_2, ..., d_l\}$ related to the topic of interest, and that each keyword in $W_0$ is contained in at least one document of $D$. %

The dictionary expansion problem can then be formulated as follows: Expand the seed dictionary $W_0$ by $m$ new words from the domain-specific corpus $D$ to obtain a data-driven \textit{expanded dictionary} 
$W=~\{w_1,...,w_n,w_{n+1},...,w_{n+m}\}$ such that 
the new words make
$W$ `more representative' of the topics of interest (compared to $W_0$) as measured by evaluation metrics such as the $F_1$ score that balances precision and recall. 

\subsection{Related work}
\textit{Keyword extraction} and \textit{query expansion} are related tasks but not equivalent to \textit{dictionary expansion}. The former refers to the extraction of representative keywords from text without an initial seed dictionary~\autocite{firoozehKeywordExtractionIssues2020}, whereas the latter relies on user input on query words or phrases, possibly expanding the current query term or suggesting additional terms \autocite{schutze2008introduction}. Both tasks have been studied mostly in the context of information retrieval in search engines and often involve relevance user feedback on retrieved documents \autocite{zheng2020bert}. 
Here, on the other hand, we use semantic relationships already captured in the latent representation space and focus on generating relevant keyword or query terms based on curated seed keywords, without explicit user feedback.

Early statistical approaches for dictionary expansion were based on ranking candidate words using %
TextRank~\autocite{mihalceaTextRankBringingOrder2004}, or by analysing word co-occurrences directly~\autocite{yinCooccurrenceBasedApproach2020}.
Promising recent work has also leveraged pre-trained word embeddings to expand a seed dictionary by choosing the most similar words according to the embeddings~\autocite{amslerUsingLexicalSemanticConcepts2020,gharibshahIKEAUnsupervisedDomainspecific2022,atteveldtCAVAOpenSource2022a,stollDevelopingIncivilityDictionary2023}.  
Prior work also suggests that global word embeddings, such as GloVe~\autocite{penningtonGloVeGlobalVectors2014}, may underperform in tasks that benefit from local properties~\autocite{diaz2016query}, which is achieved in our work through the use of local graph techniques for manifold learning and diffusion-based community detection. \cite{farmanbar2020semantic} explore domain-specific query terms but focus on important challenges in end-to-end production pipelines using direct cosine similarities. Most similar to our domain-specific setting, \cite{tsai2014financial} and \linebreak\cite{gharibshahIKEAUnsupervisedDomainspecific2022} use similar methods adapted to custom domains but do not employ the properties of the semantic network, as we do in our work.

\subsection{Motivation}\label{sec:Motivation}
A data-driven augmented dictionary can be constructed by adding words from a domain-specific vocabulary $V\subsetneq \mathbbm{R}^r$ of word embedding vectors that are most similar to the keywords in the seed dictionary $W_0\subseteq V$ according to the cosine similarity 
\begin{equation} \label{eq:cosine_similarity}
S_{\cos}(\boldsymbol{u},\boldsymbol{v}):=\frac{<\boldsymbol{u},\boldsymbol{v}>}{\|\boldsymbol{u}\|_2 \; \|\boldsymbol{v}\|_2}
\end{equation}
for word embeddings $\boldsymbol{u},\boldsymbol{v}\in V$, where $<\cdot,\cdot>$ denotes the Euclidean inner product. 
Then, given a threshold $0\le\epsilon\le1$, the \textit{thresholding-based} expanded dictionary $W(\epsilon)$ is defined as
\begin{equation}\label{eq:thresholding}
  W(\epsilon) := \bigcup_{{\boldsymbol{w}}\in W_0} \{\boldsymbol{v} \in V\;|\; S_{\cos}({\boldsymbol{w}},\boldsymbol{v})\ge\epsilon\}.
\end{equation}
Similarly, %
the \textit{kNN-expanded dictionary} $W(k)$ is defined as
\begin{equation}\label{eq:kNN-expansion}
  W(k) := \bigcup_{{\boldsymbol{w}}\in W_0} \{\boldsymbol{v} \in V\;|\; \boldsymbol{v} \in N_k(\boldsymbol{w})\},
\end{equation}
where $N_k(\boldsymbol{w})$ denotes the $k\ge 1$ most similar words to $\boldsymbol{w}$ according to  $S_{\cos}$. Increasing the parameter $\epsilon$ (or $k$) expands the seed dictionary with most similar words based on direct word similarity. To improve the quality of the expanded dictionary, the word embeddings in $V$ are typically fine-tuned on a domain-specific corpus $D$ to better capture contextual semantics for the particular task at hand. 

\begin{figure}[htb!]
  \centering
  \includegraphics[width=.9\textwidth]{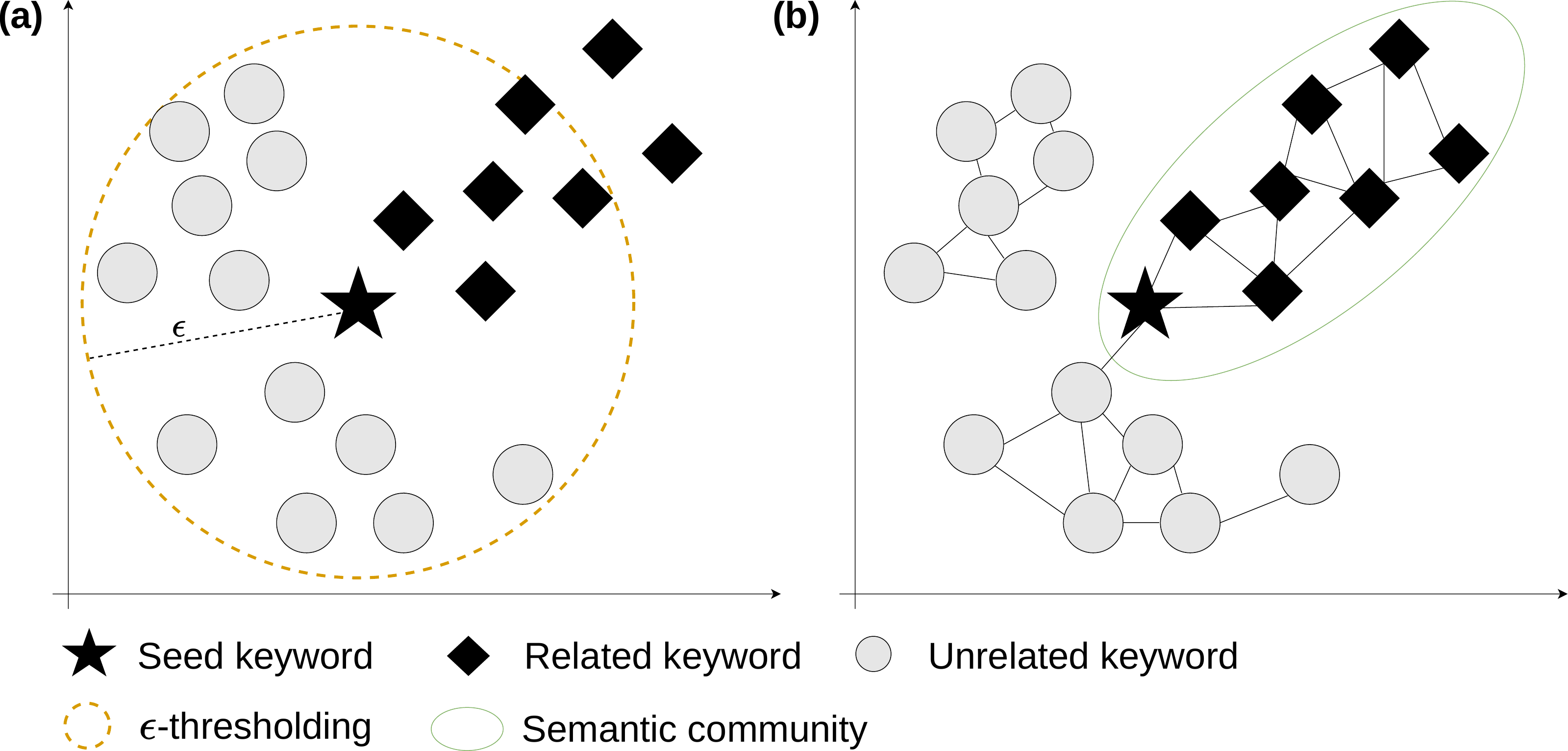}
  \caption{(a) A simple thresholding of the cosine similarity with a fixed radius $0\le\epsilon\le 1$ cannot account for inhomogeneities in the word embedding space. (b) The LGDE method based on tools from manifold learning and network science represents consistently the nonlinear local geometry of word embeddings around a seed keyword through a geometric CkNN graph and derives its semantic community through diffusion-based local community detection that captures chains of strong word associations.}
  \label{fig:LGDE_visual_abstract}
\end{figure}

An issue not easily addressed by such $\epsilon$-thresholding or kNN approaches is that direct (cosine) similarities can be uninformative in noisy, high-dimensional latent space representations, such as word embeddings. 
It is well known that the \textit{relative contrast} between closest and furthest neighbours of a set of vectors decreases rapidly for the Euclidean distance (and cosine similarity) as the dimension of the space increases~\autocite{aggarwalSurprisingBehaviorDistance2001}.
Yet, on the other hand, the high-dimensionality of word vector embeddings, typically hundreds of dimensions, is crucial to encode the rich semantic and syntactic characteristics of words learnt through training in the large datasets used in textual analysis. 
This means that there is an implicit tension when using high-dimensional word vector embeddings:  although high dimensional embeddings are desirable to enhance semantic content, high dimensionality also makes embedding vectors less distinguishable, thus 
leading  to relatively unspecific word associations when using direct word similarities.  

To help circumvent this conundrum,  LGDE uses tools from manifold learning and network science to account for %
chains of word associations and to better capture the local nonlinear geometry of seed keywords, see Figure~\ref{fig:LGDE_visual_abstract} for an illustration. 
In particular, LGDE constructs a graph (a geometric CkNN graph) to represent consistently the geometry of the local manifold of the word embeddings as a means to capture a local representation of the strongest word similarities. This graph is then explored via diffusion to obtain consistent groups of words accessible from the seed dictionary using local community detection. This construction allows us to consider chains of word associations as paths on the graph.
Indeed, a seed keyword ${\boldsymbol{w}} \in W_0$ could be similar to a word $\boldsymbol{u} \in V$ ($S_{\cos}({\boldsymbol{w}},\boldsymbol{u}) > \epsilon$) which is in turn similar to another word $\boldsymbol{v}\in V$ ($S_{\cos}(\boldsymbol{u},\boldsymbol{v}) > \epsilon$), yet we might have low direct similarity between $\boldsymbol{w}$ and $\boldsymbol{v}$  
($S_{\cos}({\boldsymbol{w}},\boldsymbol{v}) < \epsilon$), reminiscent of cosine similarity not fulfilling the standard triangle inequality~\autocite{schubertTriangleInequalityCosine2021}. 
A thresholding method based only on direct similarities would then exclude the word $\boldsymbol{v}$ from the data-driven dictionary $W(\epsilon)$, although the chain of strong word associations arguably makes $\boldsymbol{v}$ a sensible candidate for dictionary expansion. Similar problems may also occur when adding the $k$ most similar words for each seed keyword (Eq.\ref{eq:kNN-expansion}) or when thresholding is applied iteratively, e.g., with the IKEA method~\autocite{gharibshahIKEAUnsupervisedDomainspecific2022}. %

\section{Methodology}
LGDE consists of two steps. In the first step, given a set of word embedding vectors for seed and candidate keywords, we compute a similarity graph that captures the local semantic similarities of the corpus. In the second step, we use local community detection based on graph diffusion to obtain semantic communities of each seed keyword as candidates for dictionary expansion. We detail these steps in the following subsections.  We also describe a method to compute domain-specific word representations and a validation strategy for our dictionary expansion problem.

\subsection{Semantic network construction}\label{sec:cknn}
 
To explore the semantic space of our vocabulary $V\subsetneq\mathbbm{R}^r$ with size $N=|V|$, we construct an undirected, weighted \textit{semantic similarity graph} $G=(V,E)$, where the nodes correspond to the words in $V$. The weighted edges $E$ %
are computed from the $N\times N$ matrix of \textit{normalised cosine distances}:
\begin{equation}\label{eq:normalised_cosine_dist}
  \tau:=||1-S_{\cos}||_{\text{max}},
\end{equation}
where $S_{\cos}$ is the cosine similarity (Eq.~\ref{eq:cosine_similarity}) and $||\cdot ||_{\text{max}}$ is the element-wise normalisation with the max-norm so that all elements of $\tau$ are normalised between $[0,1]$. We also define the matrix of \textit{normalised cosine similarities}~\autocite{altuncuFreeTextClusters2019}:
\begin{equation}\label{eq:normalised_cosine_sim}
    S:=1-\tau.
\end{equation}

We would like to obtain a semantic network with edges weighted by the similarity $S$ but $S$ is a dense matrix that contains many small values corresponding to negligible word similarities. To produce a robust semantic network that only contains the most relevant chains of semantic similarities, we first obtain the undirected and unweighted  %
CkNN graph from the distance matrix $\tau$~\autocite{berryConsistentManifoldRepresentation2019}. The $N\times N$  adjacency matrix $B^{(k)}$ of the CkNN graph is given by:
\begin{equation}\label{eq:CkNN}
  B^{(k)}_{\boldsymbol{u},\boldsymbol{v}} := \begin{cases}
    1 & \text{if} \;\; \tau(\boldsymbol{u},\boldsymbol{v})<\delta \sqrt{\tau(\boldsymbol{u},\boldsymbol{u}_k)\tau(\boldsymbol{v},\boldsymbol{v}_k)}, \\
    0 & \text{otherwise},
  \end{cases}
\end{equation}
where $\boldsymbol{u}_k,\boldsymbol{v}_k\in V$ denote the $k$-th nearest neighbours of $\boldsymbol{u},\boldsymbol{v}\in V$, respectively, and $\delta>0$ controls the graph density. 
In contrast with a \textit{$k$-Nearest Neighbors} (kNN) graph, which does not account for inhomogeneities in the data as it connects a node to all of its $k$ nearest neighbours, the CkNN construction corrects for different densities and has been shown to approximate consistently the geometry of complex manifolds embedded in a Euclidean space~\autocite[Theorem 7]{berryConsistentManifoldRepresentation2019}.  %
Note that $\tau$ is equivalent to using Euclidean distances of normalised word vectors when $\delta=1$. In that case, and assuming  $||\boldsymbol{u}||_2=||\boldsymbol{v}||_2=1$, we have:
\begin{align*}
    &\tau(\boldsymbol{u},\boldsymbol{v})< \sqrt{\tau(\boldsymbol{u},\boldsymbol{u}_k)\tau(\boldsymbol{v},\boldsymbol{v}_k)} 
    \quad \Leftrightarrow \quad ||\boldsymbol{u}-\boldsymbol{v}||_2 <  \sqrt{||\boldsymbol{u}-\boldsymbol{u}_k||_2||\boldsymbol{v}-\boldsymbol{v}_k||_2},
\end{align*}
which follows directly from: %
\begin{align*}
    ||\boldsymbol{u}-\boldsymbol{v}||_2^2 &= ||\boldsymbol{u}||_2^2 + ||\boldsymbol{v}||_2^2 - 2\langle\boldsymbol{u},\boldsymbol{v}\rangle_2
    =2 \, \left(1-S_{\text{cos}}(\boldsymbol{u},\boldsymbol{v}) \right)=a \, \tau_{\boldsymbol{u},\boldsymbol{v}},
\end{align*}
where $a:=2\max_{\boldsymbol{u},\boldsymbol{v}\in V}(1-S_{\text{cos}}(\boldsymbol{u},\boldsymbol{v}))$ is a constant. Moreover, empirical studies have shown that CkNN with $\delta=1$ and adequate choice of $k$ outperforms other graph constructions for downstream tasks such as data clustering~\autocite{liuGraphbasedDataClustering2020} and  classification~\autocite{qianGeometricGraphsData2021}.

Finally, we can define the weighted semantic similarity network $G^{(k)}=(V,E^{(k)})$ with adjacency matrix
\begin{equation}\label{eq:weighted_semantic_network}
  A^{(k)} := S \odot B^{(k)},
\end{equation}
where $\odot$ denotes the element-wise (Hadamard) product. Therefore, the edge weights of the semantic network $G^{(k)}$ are given by the normalised semantic similarity $S$, and its backbone is the sparse CkNN graph $B^{(k)}$ that preserves the topological and local geometric features of the semantic space $V$.

\subsection{Semantic community detection}

The graph $G^{(k)}$ encodes the semantic information contained in our domain-specific corpus ${D}$ at a word-level representation with the inter-word weighted edges $E^{(k)}$ capturing relevant semantic similarities between words. Hence, paths in this graph can be interpreted as chains of word associations. Moreover, the keywords in the seed dictionary $W_0$ are nodes in the graph and we can study their context and related words by computing their local semantic community structure. 
In network science, \textit{community} refers to a group of nodes in the graph that are tightly connected within the group and more loosely connected with the other communities. \textit{Community detection} is then an unsupervised task to find such communities in graph data without specifying the number or type of communities. The result is a partitioning of the graph into mutually exclusive subgraphs (hard partitions) or into potentially overlapping subgraphs (soft partitions). See \cite{fortunatoCommunityDetectionGraphs2010} for a review and \cite{schaubManyFacetsCommunity2017,schindlerCommunityVagueOperator2023} for a discussion of the notion of community. %
Here we use the \textit{severability} method, a diffusion-based framework for fast local community detection~\autocite{yuSeverabilityMesoscaleComponents2020}, and determine the \textit{semantic community} $C(\boldsymbol{w})$ for each seed keyword $\boldsymbol{w}\in W_0$. Briefly, severability is a community detection algorithm that detects local graph communities 
by exploiting a discrete-time random walk on the graph of words, which is governed by the transition probability matrix $P$, where $P_{\boldsymbol{u},\boldsymbol{v}}$ is the probability of the random walk jumping from word $\boldsymbol{u}\in V$ to $\boldsymbol{v}\in V$:
\begin{equation}
    0\le P_{\boldsymbol{u},\boldsymbol{v}}:=\frac{A^{(k)}_{\boldsymbol{u},\boldsymbol{v}}}{\sum_{\boldsymbol{v}^\prime \in V}A^{(k)}_{\boldsymbol{u},\boldsymbol{v}^\prime}}\le 1.
\end{equation}
The semantic community $C({\boldsymbol{w}})$ of ${\boldsymbol{w}}\in W_0$ is the subset $C\subseteq V$ 
(with ${\boldsymbol{w}}\in C$) 
that maximises the severability quality function $\sigma(C,t)$ for the time scale $t $:
\begin{equation}\label{eq:severability_cost}
  0\le\sigma(C,t) := \frac{\rho(C,t)+\mu(C,t)}{2}\le 1.
\end{equation}
The severability $\sigma(C,t)$ consists of two terms: the \textit{mixing term} $\mu(C,t)$, which measures the mixing of the random walker within $C$ over time $t$, and the \textit{retention term} $\rho(C,t)$, which quantifies the probability of the random walker not escaping $C$ by time $t$~\autocite{yuSeverabilityMesoscaleComponents2020}. %
Specifically, let $Q$ denote the (substochastic) submatrix of $P$ corresponding to the nodes in a connected subset $C\subseteq V$ and let $\boldsymbol{q}_i^{(t)}$ denote the $i$-th row of the matrix $Q^t$. Then \cite{yuSeverabilityMesoscaleComponents2020} define the retention term as:
\begin{equation}
    0\le\rho(C,t):=\frac{1}{|C|}(\boldsymbol{1}^T Q^t\boldsymbol{1})\le1,
\end{equation}
where $\boldsymbol{1}$ is the $|C|$-dimensional vector of ones, and the mixing term as:
\begin{equation}
    0\le\mu(C,t):=1 - \frac{1}{|C|}\sum_{i=1}^{|C|} \left\lVert \Bar{\boldsymbol{q}} - \frac{\boldsymbol{q}_i^{(t)}}{\boldsymbol{q}_i^{(t)}\boldsymbol{1}} \right\rVert_{TV} \le1,
\end{equation}
where $\lVert \cdot \rVert_{TV}$ is the total variation distance that measures the convergence towards $\Bar{\boldsymbol{q}}$, the quasi-stationary distribution of $Q$. As $t$ increases, we need a larger-sized community $C\subseteq V$ to trap the random walk and achieve high retention $\rho(C,t)$, but simultaneously, increasing the size of $C$ makes mixing more difficult and leads to a reduced $\mu(C,t)$. 
We then optimise the severability cost function to obtain local semantic communities around the seed words based on the derived semantic similarity word graph.
To indicate the dependency of the semantic community of word $\boldsymbol{w}$ on the temporal scale $t$ and the CkNN parameter $k$ we write $C^{(k,t)}(\boldsymbol{w})$.
Note that severability captures chains of word associations through paths of length larger than two in the random walk and also allows for overlapping semantic communities (potentially capturing polysemy, if present). 
Code for optimizing the severability cost function and performing local community detection is available at \url{https://github.com/barahona-research-group/severability}.

The result of the LGDE method is then an \textit{extended dictionary} $W(k,t)$,  which is defined as the union of (overlapping) local semantic communities:
\begin{equation}\label{eq:dictionary_LGDE}
  W(k,t) := \bigcup_{\boldsymbol{w}\in W_0} C^{(k,t)}(\boldsymbol{w}).
\end{equation}
By construction $W_0\subseteq W(k,t)\subseteq V$, and the size of $W(k,t)$ generally grows with increasing $k$ and $t$ 
such that in the set-theoretic limit, we have $\lim_{k,t\rightarrow\infty}W(k,t)=V$, i.e., the extended dictionary would become the whole vocabulary. As remarked above, our extended dictionary can include words that are connected to a seed keyword $\boldsymbol{w}\in W_0$ via a chain of strong word associations, rather than only by strong direct similarity.

The computational complexity of LGDE %
depends on the vocabulary size $N$, the number of neighbours $k$ in the CkNN construction, the time scale $t$ of severability, and the size $n$ of the seed dictionary $W_0$. The complexity of the CkNN construction is $O(N^2 k)$, as it is dominated by the $\tau(\boldsymbol{u},\boldsymbol{v})$ rescaling in Eq.~\eqref{eq:CkNN}, which requires finding the $k$-th nearest neighbour (out of $N$ words) for each of the $N$ words. %
The complexity of severability is linear in $n$ with $O(n b \log_2t)$, where $b$ is a constant that depends on the maximum local community size allowed in the optimization algorithm \autocite{yuSeverabilityMesoscaleComponents2020}. 
Hence, the overall complexity for a single run of LGDE is $O(N^2 k + n b \log_2t)$. In subsequent runs, e.g., when adding a new word to the seed dictionary, no additional graph construction is required and only severability has to be applied with complexity $O(b \log_2t)$ for a single keyword. The favourable computational complexity of LGDE thus follows from the use of fast local community detection. 

\subsection{Domain-specific word representations}\label{sec:GloVe}
To construct the input space $V$, we use word embeddings generated from general-purpose corpora as vector inputs. In our setting of community detection on a word similarity graph, we wish to allow a word to belong to multiple, overlapping communities, to be able to capture multiple word meanings if they exist. Hence, we use static word embeddings such as word2vec \autocite{mikolov2013efficient} or GloVe \autocite{penningtonGloVeGlobalVectors2014} instead of contextualized embeddings such as BERT \autocite{devlin-etal-2019-bert} or ELMo \autocite{peters2018deep}.  
It may be worth noting that BERT-based~\autocite{devlin-etal-2019-bert}  models rely on subword tokenization and, typically, word embeddings are extracted by summing or averaging the subword token embeddings, a heuristic that often degrades the word-level semantic graph representation~\autocite{AltuncuPhDthesis}. This follows from the fact that BERT-style models are designed for supervised end-to-end fine-tuning, and not for extracting intermediate embedding layers.
For instance, a Semeval task on diachronic semantics \autocite{schlechtweg2020semeval} showed that static or type-based embeddings outperformed contextualized embeddings. Furthermore,  studies on representations from BERT-like models \autocite{vulic-etal-2020-probing} have shown various issues with representations produced by averaging hidden layers. These are highly task- and domain-dependent and there are no techniques to select a single layer that is reliably better \autocite{bommasani-etal-2020-interpreting}. Work on hidden layer selection for good word-level representations may be an interesting future direction that could complement our work here but is beyond the focus of the present study.

For our experiments, we use GloVe as base word embeddings, generated from general-purpose corpora like Wikipedia~2014 and Gigaword~5~\autocite{penningtonGloVeGlobalVectors2014}. The base GloVe embeddings are available in different dimensions $r=50,100,300$. It is well-known that word embeddings are dependent on the corpus they are trained on, given that terms often adopt new or additional meanings across domains or over time (e.g., ``Apple'' as a concept in technology or business domains that came into existence only in the 1970s). 
While base embeddings could be used as-is with our method, we use Mittens~\autocite{dingwallMittensExtensionGloVe2018} to tune the GloVe representations to better represent the semantic relationships in our use-case domains. In particular, for a set of $N $ words %
in our domain-specific corpus $D$, the word embeddings $\boldsymbol{v}_i$, $i=1,...,N$, are computed from a retrofitting model that optimises the Mittens cost function
\begin{equation}
\label{eq:mittens}
       J_{\text{Mittens}} = J_{\text{GloVe}} + \mu \sum_{i\in U} ||\boldsymbol{v}_i - \boldsymbol{u}_i||_2,
\end{equation}
where $J_{\text{GloVe}}$ is the standard GloVe cost function~\autocite{penningtonGloVeGlobalVectors2014} based on the word co-occurrences in %
$D$, and  $U$ %
is the index set of words for which pre-trained vector representations $\boldsymbol{u}_i$ are available. The hyperparameter $\mu \ge 0$ determines the extent of fine-tuning: a large $\mu$ favours remaining closer to the base embeddings $\boldsymbol{u}_i$, whereas a small $\mu$ favours greater adaptation of the word representations $\boldsymbol{v}_i$, such that setting $\mu=0$ means that the word vectors are computed from scratch on the new corpus. Although a value of $\mu=0.1$ was the default in \cite{dingwallMittensExtensionGloVe2018}, we find that a larger value of $\mu$ improves the quality of embeddings for a small domain-specific corpus $D$, as is the case in our study.
By training the Mittens model, we compute fine-tuned $r$-dimensional word vectors $V = \{\boldsymbol{v}_1, \boldsymbol{v}_2, ..., \boldsymbol{v}_N\}\subsetneq\mathbbm{R}^r$, for $r=50,100,300$, and we assume $W_0\subsetneq V$. Following \cite{schnabel2015evaluation} 
we normalize the word vectors to unit length, as the relative angle is sufficient to capture semantic similarity.

\subsection{Evaluation of expanded dictionaries}

Consider a domain-specific corpus of $l$ documents $D=\{d_1,...,d_l\}$ with ground-truth (human-coded) labels $\boldsymbol{y}$ such that $\boldsymbol{y}_i=1$ if document $d_i$ has a certain topic of interest (true) and $\boldsymbol{y}_i=0$ otherwise (false). To assess the quality of a data-driven dictionary $W$ we can evaluate the performance of a simple document classifier $f_W: D \rightarrow \{0,1\}$ associated with $W$, where we define $f_W(d)=1$ if there exists a keyword in $W$ that appears in the document $d\in D$ and $f_W(d)=0$ otherwise. To evaluate the trade-off between precision and recall of the dictionary $W$ on the potentially unbalanced benchmark data $(D,\boldsymbol{y})$, we compute the macro $F_1$ score of its associated classifier $f_W$ and denote this number by $F_1[W]$. Similarly, $P[W]$ denotes the macro precision and $R[W]$ the macro recall. This evaluation strategy can also be used to optimise the hyperparameters $\epsilon$, in the case of the threshold dictionary $W(\epsilon)$ in Eq.~\eqref{eq:thresholding}, or $k,t $, in the case of the LGDE dictionary $W(k,t)$ in Eq.~\eqref{eq:dictionary_LGDE}, on a train split of the benchmark data.

In addition, it is also possible to evaluate the contribution of a single keyword $\boldsymbol{w}\in W$ to the performance of the dictionary $W$. Let us consider the probability $\mathbbm{P}[w\in d_i\; |\; y_i = 1]$ that the word $\boldsymbol{w}$ appears in a true document and $\mathbbm{P}[w\in d_i\; |\; y_i = 0]$ that it appears in a false document. Then we define the likelihood ratio (LR)~\autocite{vanderhelmApplicationBayesTheorem1979} of word $\boldsymbol{w}$ as
\begin{equation}\label{eq:likelihood_ratio}
    \LR(\boldsymbol{w}):=\frac{\mathbbm{P}[w\in d_i\; |\; y_i = 1]}{\mathbbm{P}[w\in d_i\; |\; y_i = 0]},
\end{equation}
which is larger than 1 if $\boldsymbol{w}$ is more likely to appear in true documents than in false ones. Words with larger LR can thus be considered more representative of true documents in the text or corpus under study. The median LR for all words $\boldsymbol{w}\in W$ denoted by $\LR[W]$ can be used to summarise the contributions of keywords to the performance of the dictionary.

\section{Experiments}

We evaluate the performance of LGDE on two benchmark datasets and compare it to four baseline methods: (i) thresholding (Eq.~\ref{eq:thresholding}); (ii) kNN (Eq.~\ref{eq:kNN-expansion}); (iii) IKEA,  where the seed dictionary is expanded iteratively adding words with average similarity to seed and previously discovered keywords larger than a threshold until convergence~\autocite{gharibshahIKEAUnsupervisedDomainspecific2022}; and (iv) TextRank~\autocite{mihalceaTextRankBringingOrder2004}, a popular keyword extraction method that retrieves the $k$ top-ranked words according to the PageRank centrality in a word co-occurrence graph with co-occurrence window sized $w$.

\subsection{Application to a benchmark hate speech dataset}\label{sec:Redgab}

In the first experiment, we apply dictionary expansion to a seed dictionary of the most frequent hate speech-related words in a dataset collected from online platforms.

\subsubsection{Data}
We use a benchmark dataset collected by~\cite{qianBenchmarkDatasetLearning2019} from Reddit and Gab consisting of 56,099 manually annotated hate speech posts, of which 19,873 (35.4\%) are categorised as `hate speech'
according to Facebook's definition 
~\autocite[p. 4758]{qianBenchmarkDatasetLearning2019}.
The data is available at: \url{https://github.com/jing-qian/A-Benchmark-Dataset-for-Learning-to-Intervene-in-Online-Hate-Speech}.
We split the data into train data (75\%) and test data (25\%) using stratification. Following \cite{qianBenchmarkDatasetLearning2019}, we choose our seed dictionary $W_0$ as the five most frequent keywords in the benchmark dataset: ``ni**er'', ``fa**ot'', ``ret**d'', ``ret***ed'' and ``c**t''.  The task is to expand this seed dictionary with additional hate speech-related keywords.

\subsubsection{Experimental setup}\label{sec:redgab_experimental_setup}

\begin{figure*}[h!]
    \centering
    \includegraphics[width=.95\textwidth]{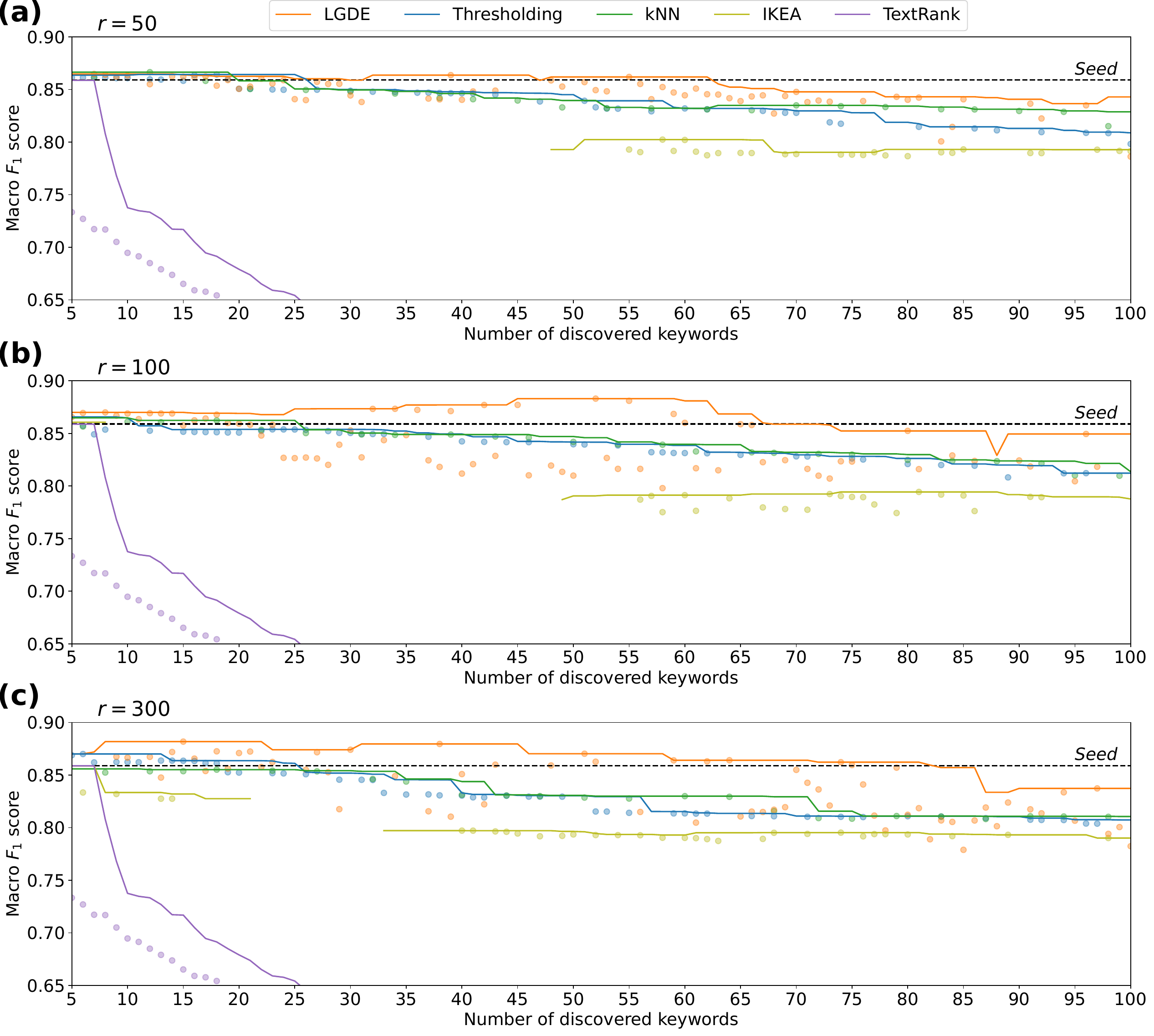}
    \caption{Hate speech benchmark dataset. Train macro $F_1$ scores for LGDE, thresholding, kNN, IKEA and TextRank dictionaries computed from word embedding dimensions (a) $r=50$, (b) $r=100$ and (c) $r=300$. Solid lines are $F_1$ scores max-pooled with window size $15$ for each method, whereas the dotted line is the train $F_1$ score of the seed dictionary for comparison. LGDE scores are consistently higher across dictionary sizes.}
\label{fig:redgab_lgde_th_comparison-300}
\end{figure*}
As outlined in Section~\ref{sec:GloVe}, we compute domain-specific word embeddings $V\subseteq\mathbbm{R}^r$, for all words appearing in at least 15 posts but in no more than $80\%$ of the posts. This selection criterion filters out stopwords, infrequent misspellings and rare words, and leads to a set with $N:=|V|=7,093$ words. We found that choosing a relatively large $\mu=1.0$ in Eq.~\eqref{eq:mittens} improves word embeddings (i.e., embeddings for out-of-vocabulary (OOV) words are learnt while retaining the quality of pre-existing word embeddings) for various embedding dimensions $r=50,100,300$. From the semantic space $V$, we compute the $N\times N$ matrix of normalised cosine similarities $S$ (Eq.~\ref{eq:normalised_cosine_sim}). 
Finally, for each dimension $r$, we compare the expansion of the seed dictionary $W_0$ into the thresholding- (Eq.~\ref{eq:thresholding}) , kNN- (Eq.~\ref{eq:kNN-expansion}), IKEA- and TextRank-based expanded dictionaries %
against the LGDE dictionary $W(k,t)$  (Eq.~\ref{eq:dictionary_LGDE}). 
For this dictionary expansion task, TextRank is only applied to documents in the training corpus with at least one seed keyword and we consider context windows $2 \leq w \leq 10$. %

As shown in Figure~\ref{fig:redgab_lgde_th_comparison-300}, LGDE consistently outperforms the other baseline methods across different sizes of the expanded dictionary. To facilitate proper comparisons, and to control for the size of the different dictionary expansion methods, we require that expanded dictionaries include at least 30 keywords but no more than 50 (with the exception of IKEA where the maximum is set to 60 due to the large gap in attained dictionary sizes). %
We then optimise the hyperparameters %
of LGDE and the other baseline methods on the train split of the benchmark data (see Table~\ref{Tab:optimal_hyperparameters} for optimal hyperparameters and Figure~\ref{fig:redgab_lgde_tuning} in the Appendix for additional information).

\begin{table*}[htb!]
    \centering
    \footnotesize
    \begin{tabular}[t]{|l||l|l|l|l|l|}
    \hline
    & LGDE & Thresholding & kNN & IKEA & TextRank\\
    \hline
    \hline
    Hate speech, $r=50$& $k=13, t=2$ & $\epsilon=0.774$ &$k=7$&$\epsilon=0.701$&$w=2, k=35$\\
    \hline
    Hate speech, $r=100$& $k=12, t=5$ & $\epsilon=0.681$ &$k=8$&$\epsilon=0.610$&$w=2,k=35$ \\
    \hline
    Hate speech, $r=300$& $k=19, t=2$ & $\epsilon=0.552$ &$k=8$&$\epsilon=0.507$&$w=2,k=35$ \\
    \hline \hline
    20 Newsgroups, $r=50$& $k=3, t=8$ & $\epsilon=0.891$ &$k=7$& - &$w=2,k=106$ \\
    \hline
    20 Newsgroups, $r=100$& $k=5, t=8$ & $\epsilon=0.807$ &$k=7$& - &$w=2,k=106$ \\
    \hline
    20 Newsgroups, $r=300$& $k=6, t=6$ & $\epsilon=0.673$ &$k=8$& - &$w=2,k=106$ \\
    \hline
    \end{tabular}
    \caption{Optimal hyperparameters for LGDE and four baselines in hate speech and 20 Newsgroups %
    experiments. In the 20 Newsgroups experiment, no hyperparameters were found for IKEA that could produce a dictionary within our size constraints (100-150 discovered words).}
    \label{Tab:optimal_hyperparameters}
\end{table*}

\subsubsection{Results}
Table~\ref{Tab:redgab_F1} shows that the optimal LGDE dictionaries $W(k,t)$ outperform the optimal %
dictionaries %
obtained with four baseline methods for all dimensions $r$ of the word embeddings, 
as measured by increased $F_1$ scores.
The quality of the LGDE dictionary improves with increasing embedding dimension and the overall best dictionary is achieved with LGDE at dimension $r=300$. On the other hand, the dictionaries based on thresholding, kNN, IKEA and TextRank do not improve upon the performance of the seed dictionary $W_0$, even when increasing the embedding dimension because of the inclusion of many non-relevant words.

\begin{table}[htb!]
    \centering
    \begin{tabular}[t]{|l||l|l|l|l|l|l|}
     \hline
     $r$ & Seed dictionary & LGDE & Thresholding & kNN & IKEA & TextRank\\
        \hline
        \hline
        $50$  & 0.856  & \textbf{0.861}   & 0.852 & 0.850 & 0.800 &   0.607 \\
        $100$ & 0.856  &\textbf{0.874}   & 0.850 & 0.848 & 0.787 &   0.607  \\
        $300$ & 0.856  &\textbf{0.875}   & 0.846 & 0.846 & 0.794&     0.607 \\
        \hline
    \end{tabular}
    \caption{Hate speech benchmark dataset. Test macro $F_1$ scores of the seed dictionary and expanded dictionaries obtained with LGDE and four baseline methods (thresholding, kNN, IKEA,  TextRank) for different embedding dimensions $r$.}
    \label{Tab:redgab_F1}
\end{table}

\begin{table*}[htb!]
    \centering
    \begin{tabular}[t]{|l||p{4cm}|p{3cm}|p{4cm}|}
    \hline
    \centering{$r$} &\centering{LGDE only}
    & \centering{Intersection}
    & \makecell{Thresholding only}\\
        \hline
        \hline
        $50$  & \footnotesize altright, \textbf{awww}, baiting, \textbf{be**er}, \textbf{braindead}, \textbf{brainless}, btw, \textbf{commie}, \textbf{commies}, \textbf{c**ts}, \textbf{f*g}, \textbf{fa***ts}, \textbf{f**ker}, gg, \textbf{goyim}, \textbf{h*g}, hater, liar, limey, mo*on, \textbf{pedo}, simp, \textbf{sk**k}, \textbf{s**ts}, snowflake, tl, \textbf{tr**ny}, \textbf{wetback}, woah 
        & \footnotesize as***le, \textbf{aww}, crybaby, \textbf{ni**a}, \textbf{ni**ers}, \textbf{prob}, sc***ag, slur, \textbf{spouting}, weasel  
        & \footnotesize \textbf{autistic}, \textbf{b**ch}, bu****it, cared, competent, convicted, \textbf{f**k}, insane, \textbf{intellectually}, \textbf{liking}, \textbf{mentally}, morally, offenders, olds, \textbf{paki}, puke, \textbf{ret**ds}, sane, ugh, yikes  \\ 
        \hline
        $100$  & \footnotesize as***les, \textbf{awww}, \textbf{badass}, \textbf{be**er}, betas, \textbf{b**bs}, btw, \textbf{chi***an}, \textbf{commies}, \textbf{c**ts}, \textbf{douche}, \textbf{fa***ts}, \textbf{f**kers}, \textbf{f**ks}, \textbf{goyim}, \textbf{ho*os}, limey, \textbf{maggot}, \textbf{motherf**ker}, motherf**kers, \textbf{pedo}, \textbf{pilled}, \textbf{prob}, \textbf{ra**ead}, shitty, simp, \textbf{sk**k}, spangle, \textbf{stfu}, th*t, \textbf{tw*t}, ummm, walkaway, \textbf{wi**er}, woah  
        & \footnotesize \textbf{b**ch}, \textbf{braindead}, \textbf{commie}, \textbf{f*g}, \textbf{f**kin}, \textbf{ni**a}, \textbf{ret**ds}, shite, \textbf{turd}, \textbf{wetback}  & \footnotesize as***le, \textbf{autistic}, cared, competent, crybaby, engineers, \textbf{f**k}, \textbf{f**king}, hillbilly, honest, incompetent, \textbf{intellectually}, \textbf{mentally}, morally, \textbf{murderers}, \textbf{ni**ers}, offenders, slur, weasel, \textbf{w**re}  \\ 
        \hline
        $300$ & \footnotesize \textbf{be**er}, \textbf{braindead}, brainwashing, btw, \textbf{commies}, \textbf{c**ts}, \textbf{douche}, \textbf{fa***ts}, foxnews, \textbf{f**ker}, \textbf{f**kers}, goalposts, \textbf{goyim}, \textbf{ho*os}, misandry, \textbf{motherf**ker}, motherf**kers, \textbf{pedo}, potus, \textbf{ra**ead}, sc***ag, shite, \textbf{sk**k}, \textbf{stfu}, sucks, th*t, \textbf{tr**ny}, \textbf{tw*t}, ummm, \textbf{wetback}, \textbf{wi**er}, woah  
        & \footnotesize \textbf{aww}, \textbf{awww}, \textbf{ni**a}, \textbf{ret**dation}, \textbf{ret**ds}, simp 
        & \footnotesize ahhh, \textbf{autistic}, \textbf{bi**h}, cared, competent, convicted, cruel, derogatory, disabled, \textbf{execute}, executed, execution, \textbf{f*g}, \textbf{f**k}, \textbf{f**king}, incompetent, mental, \textbf{mentally}, mildly, \textbf{murderers}, \textbf{ni**ers}, offenders, slur, ugh, word  \\
        
        \hline
        
    \end{tabular}
    \caption{Hate speech benchmark dataset. LGDE- and thresholding-based discovered dictionaries for different embedding dimensions $r$. Words in bold have a large likelihood ratio with $\LR[\boldsymbol{w}]\ge 2$.} 
    \label{Tab:redgab_words}
\end{table*}

A qualitative assessment of the discovered words 
(Table~\ref{Tab:redgab_words}) shows that LGDE discovers keywords relevant to hate speech documents, whereas thresholding often produces expected offensive, but largely standard, words without bringing in new semantic context. Examples of relevant keywords only found by LGDE include: ``be**er'', a racist slur for Hispanic men, ``h*g'', a misogynist slur for older women, and ``tr**ny'', a transphobic slur for a transgender people.  Such terms, including neologisms and online slang, are part of an informal jargon and potentially difficult to anticipate without being part of the relevant online community. To discover these terms using thresholding---chosen for comparison here as the best-performing baseline method---requires choosing very small values of $\epsilon$ at the price of adding irrelevant terms to $W(\epsilon)$, so that the overall $F_1$ performance is reduced. For example, discovering the term ``tr**ny'' at $r=50$ requires $\epsilon\le 0.719$ %
such that $|W(\epsilon=0.719)|=192$ with  $F_1=0.767$, as compared to the optimal LGDE score $F_1=0.861$, and at $r=300$ it requires $\epsilon\le0.485$ %
such that $|W(\epsilon=0.485)|=263$ with  $F_1 =0.741$, as compared to the optimal LGDE score $F_1=0.875$ (Table~\ref{Tab:redgab_F1}).

\begin{table}[htb!]
    \centering
    \footnotesize
    \begin{tabular}[t]{|l||ll||ll||ll||ll||ll||}
    \hline
      r & LGDE & Thresholding & LGDE & kNN & LGDE & IKEA & LGDE & TextRank\\
        \hline
        \hline
        $50$   & \textbf{2.25}  (*) & 1.45 & \textbf{2.23}  (ns) & 1.91 & \textbf{2.28}  (*) & 1.41 & \textbf{2.28}  (****) & 1.00  \\
        $100$  &\textbf{2.59} (*)  & 1.53& \textbf{2.59} (**) & 1.46 & \textbf{2.73} (***) & 1.44 & \textbf{2.79} (****) & 1.03    \\
        $300$  &\textbf{2.66}  (*) & 1.25 & \textbf{2.73}  (ns) & 1.55  & \textbf{2.97}  (**) & 1.44 & \textbf{2.97}  (****) & 0.98     \\
        \hline
    \end{tabular}
    \caption{Hate speech benchmark dataset. 
Pairwise comparisons of
median likelihood ratio LR (Eq.~\ref{eq:likelihood_ratio}) of words discovered only by LGDE {\textit vs.}  median LR of words discovered only by each of the baseline methods (thresholding, kNN, IKEA, TextRank) for different embedding dimensions, $r$.
    The asterisks indicate that the median LR of LGDE-only words is significantly higher than baseline methods-only words (ns means $p\ge 0.05$; * means $p<0.05$; ** means $p<0.01$; *** means $p<0.001$; **** means $p<0.0001$; Mann-Whitney U test). 
    } 
    \label{Tab:redgab_likelihood_ratio}
\end{table}

To further evaluate these differences, 
Table~\ref{Tab:redgab_likelihood_ratio} shows that the median likelihood ratio (LR) for words discovered only by LGDE is significantly higher than the median LR for words discovered only by thresholding, the best-performing baseline method: 
$\LR[W(k,t)\setminus W(\epsilon)]>\LR[W(\epsilon)\setminus W(k,t)]$. This result is statistically significant for all dimensions ($p<0.05$, Mann-Whitney U test) indicating that LGDE discovers words more representative of hate speech-related content. Similarly, we find that the median LR for words discovered only by LGDE is higher (and almost always significantly so) than the median LR for words discovered by each of the other baseline methods kNN, IKEA and TextRank 
(Table~\ref{Tab:redgab_likelihood_ratio}).

\subsection{Application to a benchmark 20 Newsgroups dataset}

To demonstrate the broader applicability of LGDE as a dictionary expansion method, we extend our evaluation to the 20 Newsgroup dataset in a second experiment.

\subsubsection{Data}

We use the popular \texttt{20 Newsgroups} dataset~\autocite{mitchellTwentyNewsgroups1997} comprising 18,846 posts from 20 newsgroups on different topics split into train (60\%) and test data (40\%). Following \cite{keerthiModifiedFiniteNewton2005}, we create a size-balanced two-class variant of the dataset with the positive class consisting of the 10 newsgroups with names from \texttt{sci.*}, \texttt{comp.*}, or \texttt{misc.forsale} and the negative class consisting of the other 10 groups with names from \texttt{alt.atheism}, \texttt{rec}, \texttt{soc.religion.christian} and \texttt{talk.politics}. As our seed dictionary $W_0$ we choose the 16 words that appear in the group names of the groups in the positive class: ``computer'', ``cryptography'', ``electronics'', ``graphics'', ``hardware'', ``ibm'', ``mac'', ``medicine'', ``microsoft'', ``os'', ``pc'', ``sale'', ``science'', ``space'', ``sys'', ``windows''.

\subsubsection{Experimental setup}

\begin{figure*}[ht!]
    \centering
    \includegraphics[width=.95\textwidth]{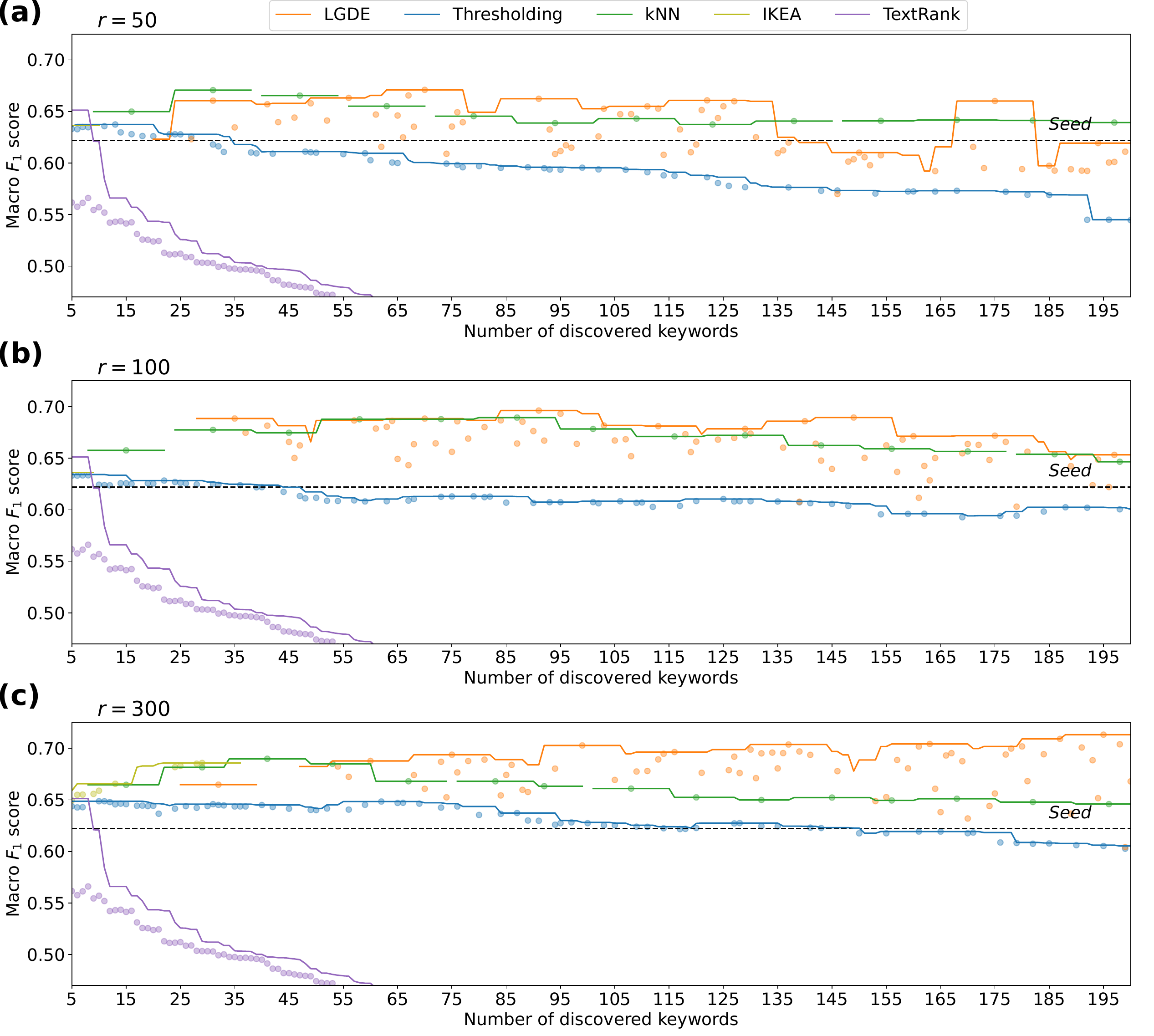}
    \caption{20 Newsgroups benchmark dataset.
    Train macro $F_1$ scores for LGDE, thresholding, kNN, and TextRank dictionaries computed from word embedding dimensions (a) $r=50$, (b) $r=100$ and (c) $r=300$. Solid lines are $F_1$ scores max-pooled with window size $15$ for each method, whereas the dotted line is the train $F_1$ score of the seed dictionary for comparison. LGDE scores are consistently higher across dictionary sizes. Note that the IKEA method does not produce dictionaries with sizes between 100 and 150 and is thus excluded from this comparison.
}
    \label{fig:newsgroups-baseline-train-comparison}
\end{figure*}

We compute domain-specific GloVe word embeddings $V\subseteq\mathbbm{R}^r$, for the words appearing in at least 15 posts and %
filter out stopwords
to obtain a set of $N:=|V|=10,751$ words. 
Due to the more standard semantic relationships in the 20 Newsgroups dataset, we chose $\mu=5.0$ in Eq.~\eqref{eq:mittens} for various embedding dimensions $r=50,100,300$. %
We then repeat the previous experimental setup for dictionary expansion as described in Section~\ref{sec:redgab_experimental_setup}. Again, we find that LGDE consistently outperforms the other baseline methods across different sizes of the expanded dictionary 
(Figure~\ref{fig:newsgroups-baseline-train-comparison}). Reflecting the larger seed dictionary size in this application, we require that expanded dictionaries include at least 100 keywords but no more than 150. Optimal hyperparameters for the different methods can be found in Table~\ref{Tab:optimal_hyperparameters}, and Figure~\ref{fig:newsgroups_lgde_tuning} in the Appendix contains additional information. Note that the IKEA method only produced dictionaries of size smaller than 30 or larger than 800 (see Figure~\ref{fig:newsgroups-baseline-train-comparison}) which were not deemed reasonable for the use case. Hence, we exclude IKEA from further comparison in this example. %

\subsubsection{Results}

Table~\ref{Tab:newsgroups_F1} shows that the optimal LGDE dictionaries $W(k,t)$ outperform the other baseline methods across all dimensions $r$ as measured by increased $F_1$ scores. Again, the performance of LGDE improves with increased dimension and the overall best performance is achieved with LGDE at dimension $r=300$. Again we find that the thresholding-based dictionaries do not improve upon the performance of the bare seed dictionary $W_0$. The kNN-based dictionaries can improve upon $W_0$ but still lead to lower performance than LGDE.

\begin{table}[htb!]
    \centering
    \begin{tabular}[t]{|l||l|l|l|l|l|}
     \hline
     $r$ & Seed dictionary & LGDE & Thresholding & kNN &  TextRank\\
        \hline
        \hline
        $50$  & 0.621  & \textbf{0.653}   & 0.580 & 0.636 &    0.437 \\
        $100$ & 0.621  &\textbf{0.692}   & 0.587 & 0.674  &   0.437  \\
        $300$ & 0.621 &\textbf{0.694}   & 0.618 & 0.649  &     0.437 \\
        \hline
    \end{tabular}
    \caption{20 Newsgroups benchmark dataset.  Test macro $F_1$ scores of the seed dictionary and expanded dictionaries obtained with LGDE and  baseline methods (thresholding, kNN,  TextRank) for different embedding dimensions $r$.
    }
    \label{Tab:newsgroups_F1}
\end{table}

We also find that words discovered only by LGDE have a significantly higher LR than words discovered only by thresholding across all dimensions (Table~\ref{Tab:newsgroups_likelihood_ratio}). %
Relative to TextRank, LGDE leads to a significantly higher LR at higher dimensions $r=100, 300$, 
whereas relative to kNN, LGDE only leads to a significantly higher LR 
at dimension $r=300$ ($p<0.01$) where the advantages of LGDE 
are more apparent. %
Overall, these results demonstrate that LGDE is an effective method of discovering relevant words for expansion above baseline methods, particularly at higher embedding dimensions that capture more semantic information from text.

\begin{table}[htb!]
    \centering
    \begin{tabular}[t]{|l||ll||ll||ll||ll||}
    \hline
      $r$ & LGDE & Thresholding & LGDE & kNN &  LGDE & TextRank\\
        \hline
        \hline
        $50$   & \textbf{2.15}  (**) & 0.92 & 1.61  & \textbf{1.99} (*) & \textbf{2.02}  (ns) & 1.17    \\
        $100$  & \textbf{4.98} (****)  & 1.47 & \textbf{3.98} (ns)  & 2.57  & \textbf{4.60} (****)  & 1.17        \\
        $300$  & \textbf{11.95}  (****) & 1.33  & \textbf{9.20}  (**) & 2.53  & \textbf{10.12}  (****) & 1.15    \\
        \hline
    \end{tabular}
    \caption{
    20 Newsgroups benchmark dataset. 
Pairwise comparisons of
median likelihood ratio LR of words discovered only by LGDE {\textit vs.}  median LR of words discovered only by baseline methods (thresholding, kNN, TextRank) for different embedding dimensions, $r$.
Statistical significance according to Mann-Whitney U test is indicated by asterisks, as in Table~\ref{Tab:redgab_likelihood_ratio}.
 } 
    \label{Tab:newsgroups_likelihood_ratio}
\end{table}

\begin{figure*}[ht!]
    \centering
    \includegraphics[width=.9\textwidth]{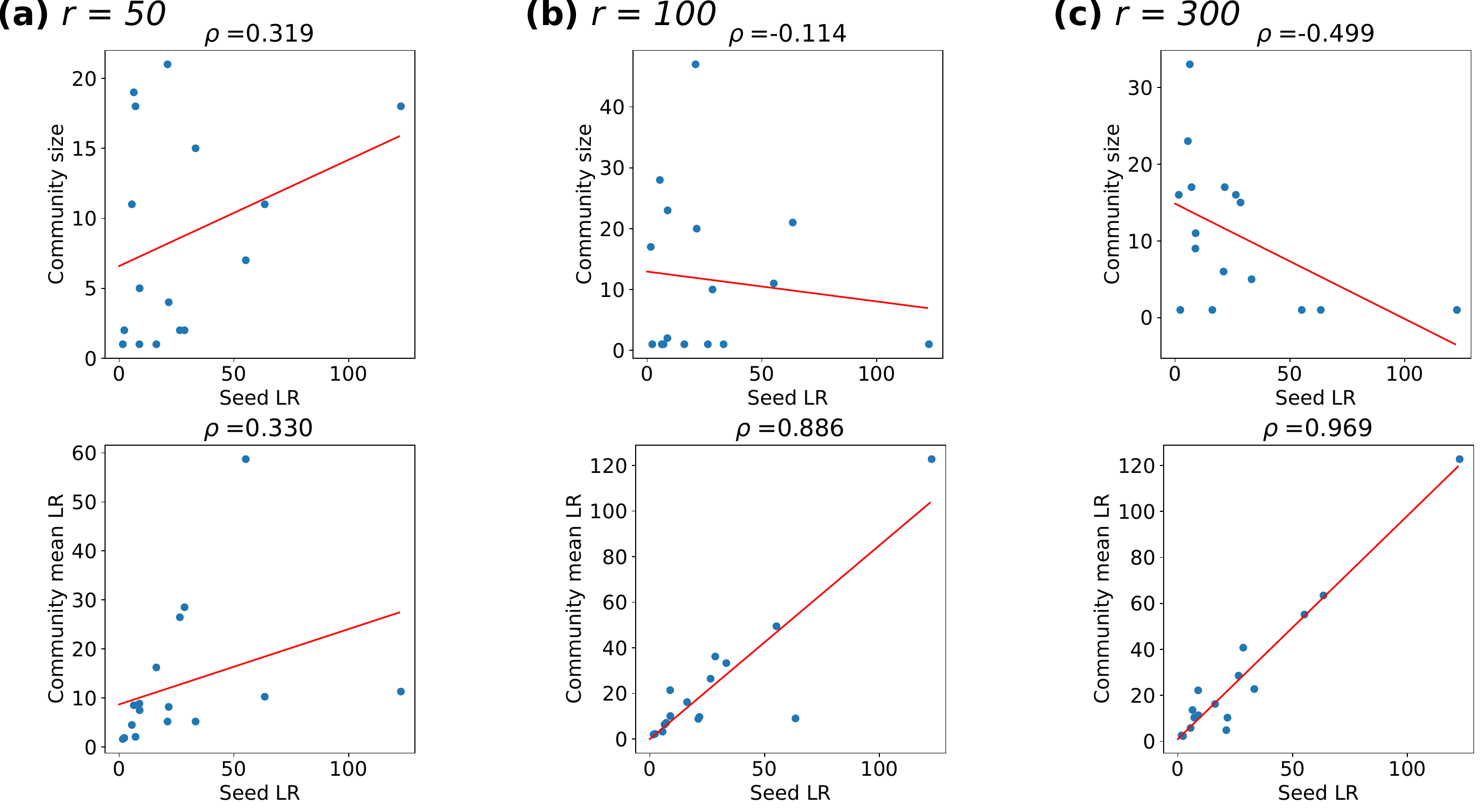}
    \caption{20 Newsgroups benchmark dataset.  Scatter plot and Pearson correlation ($\rho$) between the likelihood ratio LR of the 16 seed keywords and the size (top) or mean LR (bottom)  of their LGDE semantic community for different embedding dimensions: (a) $r=50$, (b) $r=100$ and (c) $r=300$. Red solid lines are least-squares fits. As $r$ increases, the correlation between the LR of the seed word and the size of its community becomes increasingly negative, whereas the correlation between the LR of the seed word and the mean LR of its community  becomes increasingly positive. 
    Hence, for larger embedding dimensions,  LGDE finds smaller but more specific semantic communities for high-quality seed keywords.}
    \label{fig:newsgroups-seed-corr}
\end{figure*}

Finally, in Figure~\ref{fig:newsgroups-seed-corr} we examine the relationship between the quality of the seed keyword and the quality and size of its semantic community. We find a positive Pearson correlation ($\rho$) between the LR of the seed keyword and the mean LR of the words in their semantic community, which becomes stronger as the embedding dimension grows---it increases from 0.330 for $r=50$ to 0.960 for $r=300$. On the other hand, as $r$ grows, there is also an increasingly negative Pearson correlation between the LR of the seed keyword and the size of their semantic community---$\rho$ becomes negative with value $-0.114$ for $r=100$ and decreases further to $-0.499$ for $r=300$. This means that high quality seed keywords tend to retrieve a small number of high-performing discovered words, whereas lower quality  keywords generate larger, more generic semantic communities with lower-performing discovered keywords. Note that the seed dictionary in our 20 Newsgroups experiment used solely words found in the class labels; our analysis thus highlights the importance of carefully selecting the seed dictionary.

\begin{table}[htb!]
    \centering
    \footnotesize
  (a) Hate speech benchmark dataset
    \begin{tabular}[t]{|l||l|c|c|c|c|}
     \hline
     $r$ & Seed  & LGDE (CkNN +  & CkNN + & CkNN + & CkNN +  \\
     & dictionary & severability) & SIWO & Mod $R$ & Mod $M$
     \\
        \hline
        \hline
        $50$  & 0.856  & \textbf{0.861}   & 0.859 & 0.8272 & 0.835\\
        &&&& (0.8270--0.8275) & (0.833--0.836)\\
        $100$ & 0.856  &\textbf{0.874}   & 0.827 & 0.8505 & 0.8597 \\
        &&&& (0.8493--0.8516) & (0.8594--0.8600)\\
        $300$ & 0.856  &\textbf{0.875}   & 0.828 & 0.852 & 0.862 \\
        &&&& (0.851--0.853) & (0.861--0.863)\\
        \hline
    \end{tabular}
\\ \vspace{0.3cm} 
    (b) 20 Newsgroups benchmark dataset 
    \centering
        \begin{tabular}[t]{|l||l|c|c|c|c|}
     \hline
     $r$ & Seed  & LGDE (CkNN +  & CkNN + & CkNN + & CkNN +  \\
     & dictionary & severability) & SIWO & Mod $R$ & Mod $M$
     \\
        \hline
        \hline
        $50$  & 0.621  & \textbf{0.653}   & 0.616 & 0.606  & 0.634 \\
        &&&&(0.604--0.608)&(0.632--0.637)\\
        $100$ & 0.621  &\textbf{0.692}   & 0.648 & 0.6505  & 0.659  \\
        &&&&(0.6502--0.6508)&(0.658--0.660)\\
        $300$ & 0.621  &\textbf{0.694}   & 0.664 & 0.669  & 0.663  \\
        &&&& (0.666--0.673)&(0.661--0.666)\\
        \hline
    \end{tabular}
    \caption{Test macro $F_1$ scores of the seed dictionary compared to expanded dictionaries obtained with LGDE (CkNN+severability) and CkNN with three standard local community detection methods (SIWO, local modularity $R$, local modularity $M$) for different embedding dimensions $r$:  (a) hate speech benchmark dataset; 
    (b) 20 Newsgroups benchmark dataset. Since the optimisation of local modularity $R$ and $M$ is not deterministic, we repeat the computations with 50 different seeds and present the average score and 95\% confidence intervals for dictionaries within the admissible size range.}
    \label{Tab:F1_LCD}
    \vspace{-0.3cm}
\end{table}

\section{Comparison of local community detection methods}\label{Sec:comparision_LCD}

In this section, we examine the use of severability as a local community detection method in the LGDE pipeline. As described in the introduction, 
severability complements consistently the CkNN construction of semantic networks as they are both underpinned by graph diffusions, which we use as our basis for dictionary expansion.
Whereas most community detection methods are global in nature, we have compared to popular   
 local community detection methods~\autocite{baltsouLocalCommunityDetection2022}: \textit{local modularity} $R$~\autocite{clausetFindingLocalCommunity2005}, \textit{local modularity} $M$~\autocite{luoExploringLocalCommunity2006} and \textit{SIWO}~\autocite{zafarmandFastLocalCommunity2023}.
However, these methods face several limitations in the context of dictionary expansion. Firstly,  local modularity $R$ and $M$ are defined for unweighted graphs and thus fail to capture the weighted semantic similarity in our word network. 
Secondly, the fact that the optimisation of local modularity $R$ and $M$ is not deterministic (i.e., seed dependent) introduces additional, undesired variability in the dictionary expansion task.
Thirdly, these algorithms exploit combinatorial features (density of edges for local modularity $R$ and $M$ 
or triangles for SIWO) to identify a local community, 
thus neglecting paths and the ensuing chains of strong word associations. Finally, these methods are parameter-free and return fixed-size local communities that do not accommodate naturally the different sizes of local communities. 
Our experiments confirm that severability outperforms the three other local community detection methods for our dictionary expansion task in both the benchmark hate speech dataset and the 20 Newsgroups dataset, as shown in the  
consistently higher $F_1$ scores in 
Table
\ref{Tab:F1_LCD}.

\section{Application to conspiracy-related content on 4chan}\label{sec:4chan-application}
As an application to a real-world use case, we apply LGDE to the problem of collecting conspiracy-related 4chan posts. All content on 4chan is ephemeral and together with complete user anonymity and the absence of content moderation a highly vernacular user culture has developed, partly characterised by racist and misogynous content~\autocite{dezeeuwTehInternetSerious2020}. We follow the definition of conspiracy-related content in~\cite[p. 3]{heftChallengesApproachesData2023} as ``the full public communication \textit{on} and \textit{about} (alleged) conspiracy theories in various communication venues, including their narrations, counter-narrations, and debunking, as well as neutral observation forms''. 
Detecting such content can be challenging as participants use slang and insider humour to explicitly distinguish themselves from a perceived out-group~\autocite{dezeeuwTracingNormieficationCrossplatform2020}. Therefore, terms used in public debate or the scientific literature might deviate from the vocabulary used by 4chan users. Furthermore, the vocabulary and the narratives change over time as new events lead to adaptations of existing narratives or new theories emerge and are then included in the existing canon~\autocite{garryQAnonConspiracyTheory2021, heftChallengesApproachesData2023}.
Therefore, relying solely on pre-existing domain knowledge in such an open-ended, evolving domain often leads to high-precision but low-recall retrieval methods. In particular, %
using only literature-based dictionaries to retrieve conspiracy-related posts from 4chan 
could miss neologisms and creative composition of existing concepts, leading to incomplete collections of longitudinal datasets of relevant user posts. Improving the recall is thus crucial for data collection in communication science and requires improved methods of dictionary expansion. Below, we show that starting from an expert-selected seed dictionary, LGDE discovers new conspiracy-related words that would be missed without a local graph-based perspective. %

\subsection{Data}\label{4chan-data}
We assemble an \textit{initial} seed dictionary $\widetilde{W}_0$  with 215 keywords representative of two conspiracy theories (`Great Replacement'
and `New World Order' 
) 
based on the RPC-Lex dictionary~\autocite{puschmannRPCLexDictionaryMeasure2022} and other literature  (full list in Table~\ref{tab:4chan_seed_dict}). %
Using the \mbox{fouRplebsAPI}~\autocite{buehlingFouRplebsAPIPackageAccessing2022}, we collect all English language posts published in 22 sample weeks (2 weeks in each year from 2011 to 2021, see Table~\ref{Tab:4chan_sample_weeks} in Appendix) on 4chan's political discussion board \texttt{/pol/} (\texttt{https://boards.4chan.org/pol/}) containing at least one of the words in $\widetilde{W}_0$. This leads to a %
corpus $D$ with $102,058$ unique documents. Since many conspiracy-related keywords are multi-word phrases (e.g.,  ``great replacement''), we pre-process the input to include hyphenated terms and noun phrases.
In particular, we first detect multi-word phrases in the corpus that are (1) seed keywords, (2) hyphenated terms, or (3) noun phrases with spaCy~\autocite{honnibalSpaCyIndustrialstrengthNatural2020} 
and then make the phrases uniform by lemmatizing their tokens that are nouns, %
verbs, adjectives, and adverbs. 
We then use Stanza~\autocite{qiStanzaPythonNatural2020} for further pre-processing of the data. 

We prepared two human-coded benchmark datasets as follows. As a training set, a random sample of 500 documents from $D$ was given a `yes/no' label of being conspiracy-related according to the majority vote of three independent human coders (trained student assistants). We found that 65 out of 500 documents (13.0\%) were classed as conspiracy-related. We independently collected a corpus of 47,272 4chan posts from the sample weeks. To reduce the imbalance of the dataset, we clustered document-level embedding representations of these samples using the $k$-medoids method~\autocite{parkSimpleFastAlgorithm2009}. This lets us identify 225 documents as those medoids which are most similar to, but distinct from, the posts with coded ground-truth. These 225 documents serve as our test set, of which 69 are conspiracy-related (34.5\%) according to the majority vote of the three independent human coders. To avoid extended exposure of human coders to potentially abusive 4chan content, we kept the size of the manual annotation datasets small. %

\subsection{Experimental setup}

The vocabulary $V$ is formed by the 5000 most frequent words (excluding stop words) in the domain-specific corpus $D$ and it includes our seed keywords. %
We compute fine-tuned word embeddings $V\subseteq\mathbbm{R}^{100}$ from $D$, starting from pre-trained 100-dimensional GloVe base embeddings %
using Mittens in Eq.~\eqref{eq:mittens} with the default $\mu=0.1$, as our corpus $D$ is reasonably large~\autocite{dingwallMittensExtensionGloVe2018}. 
Our seed dictionary $W_0=\Tilde{W}_0 \cap V$ is comprised of the 109 keywords that appear in the corpus.

Next, we compare the dictionary expansions obtained using thresholding 
and LGDE 
with a maximum of 150 %
discovered keywords. We perform hyperparameter tuning to obtain the optimal dictionaries by evaluating the performance of the discovered words $W(\epsilon)\setminus W_0$ and $W(k,t)\setminus W_0$, since our human-coded train data was collected using the seed dictionary $W_0$. We find that, in this case,  $k=7$ and $t=1$ leads to the best dictionary for LGDE and $\epsilon=0.730$ for thresholding. %

\subsubsection{Protocol for expert-based annotation}\label{berlin-annotation}
Three domain experts, proficient in identifying and collecting data related to conspiracy theories, were presented with the complete set (union) of words discovered by the thresholding method and LGDE, in a random order. Each expert independently carried out blind annotation of whether discovered terms are suitable to be used as a keyword to search for conspiracy-related content on 4chan, as defined above~\autocite[p. 3]{heftChallengesApproachesData2023}. The annotation guideline required a keyword to be labelled as 1 if: it is related to a narration of a specific kind of conspiracy theory; or it talks about possibly conspicuous actors and actions; or it is a form of ‘outside’ observation and reporting on actors spreading conspiracy theories or actions perceived by others as conspiracies (or 0 otherwise). The final labels are assigned based on majority voting. We discuss the results in the next section.

\subsection{Results}

We first evaluate the performance of the thresholding and LGDE expanded dictionaries using the human-coded test data described in Section \ref{4chan-data}. Table~\ref{Tab:4chan} shows that the LGDE dictionary %
has the highest macro $F_1$-Score of 0.629, and LGDE achieves both higher macro precision and recall than the thresholding-based dictionary. %
As expected, the expert-selected seed dictionary has the highest precision, $P$, but LGDE improves substantially the recall, $R$, and hence the $F_1$ score, whereas the thresholding-based expansion is not as effective.

\begin{table}[htb!]
    \centering
    \begin{tabular}[t]{|l||l|l|l|}
    \hline
      & $P$      & $R$ & $F_1$ \\
        \hline
        \hline
        Seed dictionary &\textbf{0.769}&0.559& 0.529 \\
        Thresholding &0.595&0.609&0.558       \\
        LGDE &0.700&\textbf{0.621}&\textbf{0.629}\\      
        \hline
    \end{tabular}
    \caption{Test macro scores (precision $P$, recall $R$, $F_1$ score) for the seed, threshold-expanded and LGDE dictionaries in the conspiracy-related dataset.}
    \label{Tab:4chan}
\end{table}

\begin{table*}[ht!] 
    \centering
    \begin{tabular}[t]{|p{0.02\textwidth}||p{0.9\textwidth}|}
    \hline
    \vspace{-0.3cm}\rotatebox{90}{\footnotesize LGDE only} & \footnotesize :\^\;, adolf\_hitler, afd, ain't, alt\_right, amalek, american\_government, architect, bannon, \textbf{bioweapon}, \textbf{bogdanoffs}, bombing, btfo, \textbf{cabal}, cattle, civic\_nationalism, cold\_war, communist\_party, cont, ctr, cull, death\_camp, \textbf{disguis}, donald\_trump, durr, eas, \textbf{end\_game}, end\_result, \textbf{end\_time}, engineering, \textbf{entire\_population}, etc, executive\_order, expulsion, extermination\_camp, \textbf{false\_narrative}, \textbf{false\_prophet}, general\_public, george\_washington, gib, \textbf{globalist\_jews}, \textbf{globalist\_puppet}, \textbf{globoho*o}, \textbf{golem}, \textbf{good\_goy}, good\_job, good\_people, great\_ally, hahaha, hahahaha, happ, ho*o, hurr, imo, international\_criminal\_court, invasion, \textbf{israeli\_puppet}, \textbf{jeff\_bezos}, jesus\_christ, \textbf{jewish\_plan}, \textbf{jewish\_role}, \textbf{jewish\_state}, \textbf{jp\_morgan}, \textbf{juden}, kamala\_harris, lebensraum, libt**d, main\_reason, \textbf{mass\_extinction}, \textbf{master}, memeflag, mental\_illness, \textbf{mkultra}, mutt, muzzy, new\_age, new\_zealand, nonce, \textbf{nsg}, occupy, oligarch, orthodox\_church, ottoman\_empire, pharisee, pogrom, \textbf{predictive\_programming}, prev, ptg, pu**ie, \textbf{racemixing}, \textbf{real\_power}, removal, \textbf{rule\_class}, saa, schlo*o, scientist, sperg, stooge, supreme\_court, tldr, \textbf{tpp}, vast\_majority, video\_game, welfare\_state, white\_supremacy, year\_old, \textbf{zog\_puppet} \\ 
    
    \hline 
    \vspace{-0.3cm}\rotatebox{90}{\footnotesize Intersection} & \footnotesize \textbf{9/11}, admit, \textbf{annihilation}, assimilation, atrocity, banker, be**er, billionaire, \textbf{cia}, commie, commy, conquer, degeneracy, \textbf{depopulation}, deportation, destroy, eliminate, engineer, extinct, \textbf{financier}, \textbf{freemasonry}, heroe, invade, k**e, \textbf{luciferian}, \textbf{masonic}, \textbf{mass\_murder}, massacre, neocon, ohhhh, \textbf{open\_border}, pedo, persecution, \textbf{relocate}, rid, \textbf{rockefeller}, \textbf{shill}, slaughter, species, terror, wipe, \textbf{wwiii} \\ 
    \hline 
    
    \vspace{-0.3cm}\rotatebox{90}{\footnotesize Thresholding only} &  \footnotesize 2., 3., act, annihilate, armenian, \textbf{assassinate}, \textbf{assassination}, attack, attempt, auschwitz, banish, bank, basically, biden, capitalism, capitalist, colonize, commit, communism, communist, conservatism, \textbf{conspiracy}, consumerism, crime, c**k, cult, cure, deport, \textbf{displace}, diversity, \textbf{elite}, embrace, empire, \textbf{enslave}, eu, existence, expansion, exploit, fa***t, fascism, feminism, fuck, fund, globalization, guy, happen, hate, hillary, \textbf{holocaust}, homosexuality, human, imperialism, internationalism, isolate, \textbf{jesuit}, jew, jews, justify, \textbf{k**e\_puppet}, kill, leftist, liberal, liberalism, liberate, like, literally, marxist, \textbf{mass\_genocide}, member, money, \textbf{mossad}, murder, nationalism, nazi, need, neoliberalism, ni**er, oppress, president, promote, protect, push, racism, rape, religion, rwanda, save, shit, shitskin, socialism, spread, \textbf{subjugate}, terrorist, tolerance, torture, tribunal, trump, try, uk, vatican, violence, war, weed, win, \textbf{zionism}, \textbf{zionist}  \\

        \hline
        
    \end{tabular}
    \caption{Discovered words in the 4chan dataset. Words in bold are classified as conspiracy-related by majority vote of three independent domain experts (30.2\% of the words discovered by LGDE \textit{vs.} 18.9\% of the words discovered by thresholding). LGDE discovered significantly more conspiracy-related words than thresholding ($p<0.01$, Fisher's exact test). %
    }
    \label{Tab:4chan_words}
\end{table*}

The words discovered through both dictionary expansions were evaluated and annotated according to the majority vote of three independent domain experts following the protocol described in Section \ref{berlin-annotation}. We find that LGDE discovers significantly more conspiracy-related keywords ($p<0.01$, Fisher's exact test): 30.2\% of the words discovered by LGDE are found to be conspiracy-related in contrast to 18.9\% of the words discovered by thresholding (Table~\ref{Tab:4chan_words}). 

A qualitative assessment of the discovered terms shows that this quantitative improvement coincides with the semantic content of the discovered words.  Many of the terms discovered via thresholding are formal words relating to parts of the population, political philosophies, individuals, or entities, whereas the words discovered via LGDE are more closely associated with 4chan users' platform-specific rhetoric. For example, LGDE discovers anti-Semitic jargon that might seem unremarkable in other contexts, such as ``golem'', an anti-Semitic concept that refers to people ``brainwashed'' by Jewish media, or ``good goy'', an anti-Semitic term related to Jewish domination. Other key vocabulary related to conspiracy narratives include  ``globoho*o'', an anti-Semitic and homophobic portmanteau of the words ``global'' and ``homosexual'', ``predictive programming''\footnote{\hyperlink{https://u.osu.edu/vanzandt/2018/04/18/predictive-programming/}{``Predictive programming''} refers to the idea that governments use popular media like books or films to control people.} or ``MKUltra''\footnote{\hyperlink{https://en.wikipedia.org/wiki/MKUltra}{``MKUltra''} refers to human experiments conducted by the CIA in the past including psychological torture.}. Our corpus contains noun phrases as terms, and we observe that LGDE captures more multi-word phrases useful for the identification of conspiracy-related posts, such as ``Jewish plan'' or ``Israeli puppet'', instead of the less useful unigrams ``Jewish'', ``plan'' or ``Israeli''.

\section{Discussion}

In this work, we have proposed LGDE, a graph-based, local-diffusion framework for the discovery of keywords that are semantically similar to a pre-defined seed dictionary. Using tools from manifold learning and network science, LGDE captures the complex nonlinear geometry of word embeddings and finds not only most similar words but also chains of word associations. %
Our results on two benchmarks (a standard 20 Newsgroups dataset and a hate speech corpus from online posts) show that LGDE outperforms baseline methods (thresholding, kNN, iterative thresholding (IKEA), TextRank) with improved 
macro $F_1$ scores and larger likelihood ratio for words discovered, indicating that LGDE expands the dictionary with more representative terms. Keyword expansion is especially important in highly dynamic text corpora (such as the online hate speech example), which typically employ an informal lexicon or jargon that is challenging to anticipate without in-depth knowledge of online communities.

\begin{table}[htb!]
    \footnotesize
    \centering
    \begin{tabular}[t]{|l||c|c|c|c|c|}
     \hline
      & Word & Geometric %
      & Graph & Local & Diffusion- \\
      & embeddings & cut-off
      &  & community & based \\
        \hline 
        \hline
        TextRank &   &   & $\bigstar$ &   & $\bigstar$\\
        \hline
        Thresholding & $\bigstar$ & $\bigstar$ &   &   &   \\
        \hline
        IKEA & $\bigstar$ & $\bigstar$ &   &   &   \\
        \hline
        kNN & $\bigstar$ &  & $\bigstar$  &   &   \\
        \hline
        CkNN + Mod $R$ & $\bigstar$ &   & $\bigstar$ & $\bigstar$&    \\
        \hline
        CkNN + Mod $M$ & $\bigstar$ &   & $\bigstar$ &$\bigstar$ &   \\
        \hline
        CkNN + SIWO & $\bigstar$ &   & $\bigstar$ & $\bigstar$&    \\
        \hline
        LGDE (CkNN + severability)   & $\bigstar$ &   & $\bigstar$ & $\bigstar$ & $\bigstar$ \\
        \hline
    \end{tabular}
    \caption{Overview of dictionary expansion methods considered here. LGDE is the only graph-based method applied to word embeddings that uses diffusion-based local community detection.}
    \label{Tab:methods_overview}
\end{table}

Table \ref{Tab:methods_overview} presents a comparative  overview of the principles underpinning the different methods examined here.
TextRank stands in contrast to the other methods in that it does not rely on similarities from word embeddings but rather selects relevant words by ranking them according to their PageRank, a random-walk centrality in the graph of word co-occurrences.
In our experiments, TextRank disproportionately included less relevant terms and resulted in reduced $F_1$ and LR scores, thus %
confirming the importance of going beyond mere word co-occurrences, and instead considering the rich semantic representation provided by high-dimensional word embeddings.

The rest of the methods considered here use word embeddings and exploit the geometry of the embedding space in different ways. 
In particular, both thresholding and IKEA enlarge the dictionary with neighbourhoods around seed words defined via a geometric cut-off---in one swoop in the case of thresholding, and iteratively in the case of IKEA.
Neither tresholding or IKEA improve upon the $F_1$ scores of the seed dictionaries in either the hate speech or 20 Newsgroups data, and are both outperformed by LGDE both in terms of $F_1$ scores and likelihood ratios in all cases.
In our experiments, the iterative IKEA method failed to effectively manage the trade-off between discovering new terms and maintaining the dictionary quality, and produced  
dictionaries that were either too small or excessively large with low overall performance and limited applicability for real-world use cases.
These results from thresholding and IKEA reveal the limitations of using Euclidean neighbourhoods in a high-dimensionality and highly nonlinear embedding space.

One way to capture the complexity of the embedding space is through the use of graphs to describe its local geometry, as is done in geometric and manifold learning. All the remaining methods in Table~\ref{Tab:methods_overview} use local geometric graphs to capture the neighbourhoods of seed words.  
The simplest of all is the kNN graph constructed from the $k$ closest neighbours, which shows improved $F_1$ scores relative to the seed dictionary in the 20 Newsgroups data. However, since kNN dictionary expansion only considers direct word similarities, it fails to take advantage of additional properties of the local graph construction. This leads to lower $F_1$ and likelihood ratio scores, 
especially as the dimensionality of the embedding space increases.

The remaining methods not only construct a graph representation of local neighbourhoods (via CkNN), but also exploit graph-theoretical properties to obtain local communities around each seed word.  Three of the local community detection methods (SIWO, local modularity $R$ and $M$) use combinatorial graph properties (edge or clique high density) to identify local communities as a means to capture neighbourhood inhomogeneity in embedding space. In our setting, these criteria translate into semantic communities with consistent, high word similarities, and these methods often improve on the seed dictionary and can outperform other baseline methods.
However, the reliance on density graph measures for local community detection limits the effectiveness of CkNN graph-based manifold learning for dictionary expansion. 

LGDE, which employs a CkNN graph construction combined with severability to identify local communities based on graph diffusions, consistently outperforms the other methods in our experiments. 
Using random walks to explore graph neighbourhoods in an adaptive manner allows for the identification of local communities that include strong long paths that capture semantic similarity via chains of word associations. 
LGDE can thus take advantage of the semantic content in high-dimensional word embeddings, while avoiding the lack of contrast that ensues from high dimensionality. 
This is seen in the higher $F_1$-scores and likelihood ratios for LGDE as the dimensionality of the word embeddings increases from $r=50$ to  $r=300$, in contrast to the reduction of performance observed for the thresholding-, kNN- and IKEA-based approaches. 

Although LGDE performs well in dictionary expansion for both standard text from the 20 Newsgroups and online hate speech text with highly subjective and evolving language usage, 
LGDE may be particularly suitable in cases like the latter, which require differentiation between complex topics or dealing with specialised jargon or colloquial vocabulary. In particular, we find that in the real-world use case of a corpus of conspiracy-related 4chan posts, LGDE discovers platform-specific words, noun phrases and jargon terms that represent the neologisms and informal register specific to the 4chan corpus. This makes LGDE especially informative when researchers cannot assume  \textit{a priori} a comprehensive knowledge of the lexical variety of the object of study.

\begin{figure}[htb!]
  \centering
  \includegraphics[width=\textwidth]{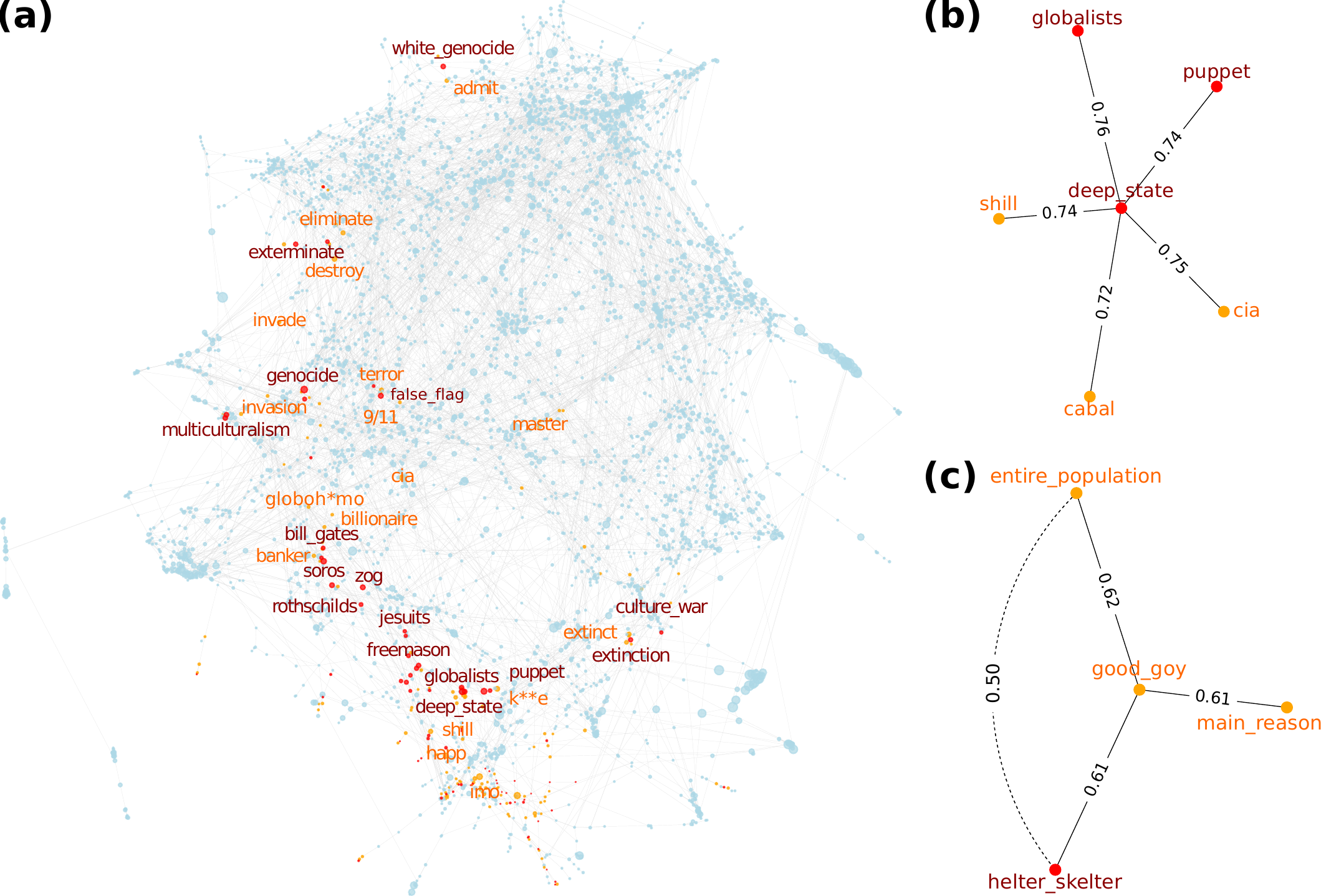}
  \caption{(a) Largest connected component of the semantic similarity graph obtained within LGDE for the 4chan dataset, where red nodes indicate seed keywords and orange nodes correspond to discovered words. (b) The semantic community of ``globalists'' contains two other seed keywords and three discovered words, all conspiracy-related according to domain expert evaluation, see Table~\ref{Tab:4chan_words}. The edge weights represent word similarities. (c)~The semantic community of ``helter\_skelter'' contains the two conspiracy-related discovered words ``good\_goy'' and ``entire\_population'', where the latter is connected to the seed keyword via a chain of high similarities whereas the direct similarity (0.50) to the seed keyword (dashed line) is lower. If such a low similarity is used in the thresholding-based approach, the expanded dictionary $W(\epsilon)$ would include 4891 words (out of 5000 in the whole vocabulary) with a low performance of $F_1=0.246$ on the test set. }
  \label{fig:4chan_network}
\end{figure}

As a final illustration of LGDE,  Figure~\ref{fig:4chan_network} shows the semantic similarity graph for the 4chan dataset with the conspiracy-related seed keywords (red)  and discovered words (orange) clustered closely together on the left side of the layout. Semantic communities obtained by LGDE often overlap and can contain multiple seed keywords, e.g., the semantic community of ``deep state'' in Figure~\ref{fig:4chan_network} (b). The semantic communities can also contain chains of strong word associations.
For example, the seed word ``helter\_skelter''\footnote{\hyperlink{https://en.wikipedia.org/wiki/Helter_Skelter_(scenario)}{``Helter\_skelter''} refers to an apocalyptic vision of race conflict.} is linked to the word ``entire\_population'', a term  
used within `Great Replacement narratives, via a chain through the term ``good\_goy'' with similarities of 0.61 and 0.62, respectively, whereas the direct similarity is only 0.50, see  
Figure~\ref{fig:4chan_network} (c). If we were to set $\epsilon=0.50$ in a thresholding approach to discover ``entire\_population'' from the seed ``helter\_skelter'', the expanded dictionary would contain 4891 words (almost the full vocabulary) with a poor test $F_1=0.246$. 
In summary, the improved performance of LGDE stems from the combination of a graph-based description of the local geometry of the word-embedding space together with the diffusion-based detection of semantic communities in the local neighbourhoods of seed words.

\section{Limitations and future work}

We list some limitations in the current work and areas of open research on which we would like to expand in future. While our experiments here employ only English language data, our method is general and the application could be useful for similar data in other languages.  Also, although qualitative assessment is invaluable for the validation of our results, the process of manual annotation can be slow and costly and, in particular, %
labelling hate speech- or conspiracy-related content can pose severe mental health risks to human annotators. Hence, in future work, we would like to use LGDE as part of mixed methods approaches ~\autocite{puschmannRPCLexDictionaryMeasure2022}. 
This is especially important for domains like hate speech or conspiracy-related communication where systems based on Large Language Models (LLMs)~\autocite{jagermanQueryExpansionPrompting2023, wangQuery2docQueryExpansion2023,leiCorpusSteeredQueryExpansion2024} could be curtailed due to strict moderation filters.
Other areas of further work include other specialised domains where the adaptation, creation and evolution of terminology is a central feature, e.g., in the scientific literature in relation to the emergence of new sub-disciplines or research areas. Furthermore, it will be important in future to carry out a detailed error analysis of LGDE to better understand the limitations of the method.
It would also be interesting to evaluate the applicability of LGDE to specialised word disambiguation tasks, since we have observed preliminary evidence of polysemy being captured through overlapping semantic communities~\autocite{yuSeverabilityMesoscaleComponents2020}.

\section*{Data and code availability}%
The hate speech data is publicly available at \url{https://github.com/jing-qian/A-Benchmark-Dataset-for-Learning-to-Intervene-in-Online-Hate-Speech} 
and the 4chan data may be made available upon evaluation of reasonable requests. Code for optimizing the severability cost function and performing overlapping community detection is available at \url{https://github.com/barahona-research-group/severability}. An implementation of LGDE and code to reproduce the results and figures in this study is available at \url{https://github.com/barahona-research-group/LGDE}.

\section*{Funding}%
Juni Schindler acknowledges support from the EPSRC (PhD studentship through the Department of Mathematics at Imperial College London) and the Weizenbaum Institute (Research Fellowship). Annett Heft, Kilian Buehling and Xixuan Zhang acknowledge support by grants from the German Federal Ministry of Education and Research (grant numbers 13N16049 [in the context of the call for proposals Civil Security – Societies in Transition] and 16DII135 [in the context of the Weizenbaum Institute]). Mauricio Barahona acknowledges support from EPSRC grant EP/N014529/1 supporting the EPSRC Centre for Mathematics of Precision Healthcare.

\section*{Acknowledgments}
This study has benefited from discussions in the \textit{Advancing Cross-Platform Research in Political Social Media Communication} workshop at the Weizenbaum Institute in 2022 and from the first author's presentation in the Computational Methods Division at the International Communication Association Conference in 2023. We thank our student assistants Joana Becker, Dominik Hokamp, Angelika Juhász, and Katharina Sawade for their invaluable work of labelling benchmark data of conspiracy-related 4chan posts. We also thank Yun William Yu and Angad Khurana for helpful discussions on the implementation of severability, and Asem Alaa for valuable suggestions on the Python packaging of the code for this project. Finally, we are grateful to the anonymous reviewers for their constructive comments that significantly enhanced the manuscript.
\appendix
\section{Appendix}

\begin{figure}[H]
    \centering
    \includegraphics[width=\textwidth]{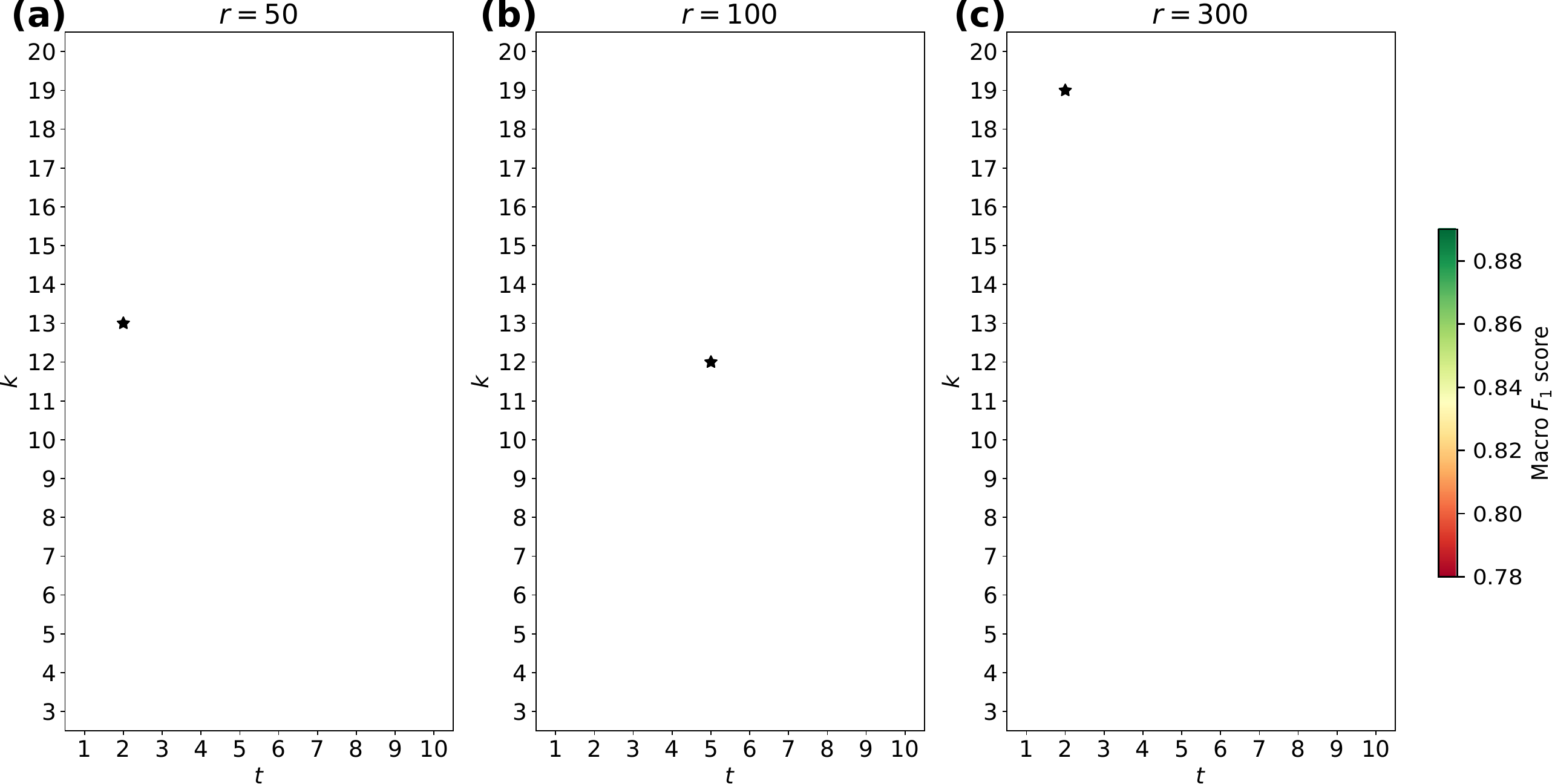}
    \caption{Hyperparameter tuning of $(k,t)$ for LGDE dictionary $W(k,t)$ in the hate speech dataset is performed using the train macro $F_1$ score of expanded dictionaries (optimal values indicated by stars). Note that the hyperparameter combinations where the number of discovered words is less than 30 or more than 50 are not considered (shaded fields).}%
    \label{fig:redgab_lgde_tuning}
\end{figure}

\begin{figure}[H]
    \centering
    
    \includegraphics[width=\textwidth]{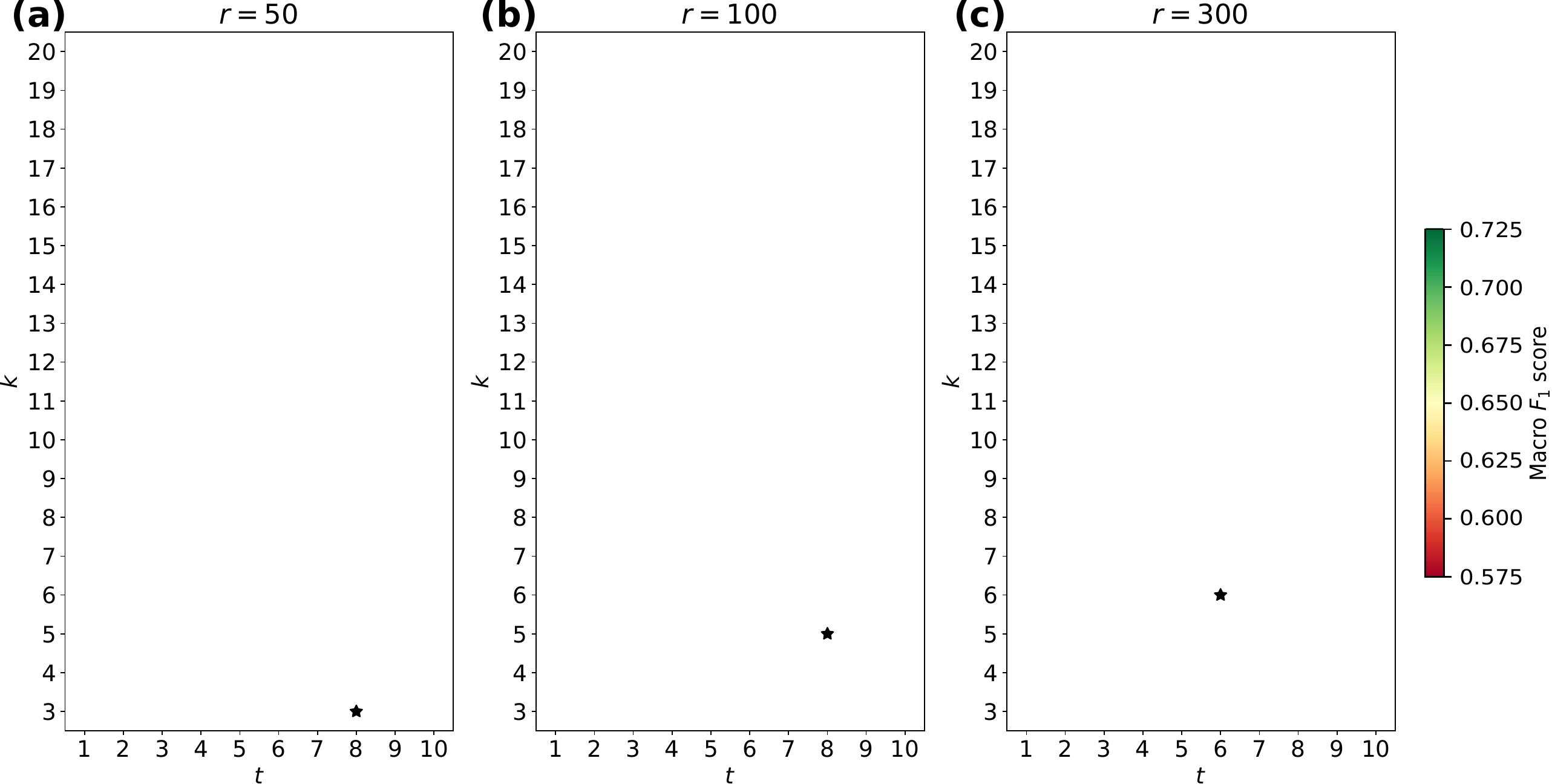}
    \caption{Hyperparameter tuning of $(k,t)$ for LGDE dictionary $W(k,t)$ in the20 Newsgroups dataset is performed using the train macro $F_1$ score of expanded dictionaries (optimal values indicated by stars). Note that the hyperparameter combinations where the number of discovered words is less than 100 or more than 150 are not considered (shaded fields).}%
    \label{fig:newsgroups_lgde_tuning}
\end{figure}

\begin{table}[H]
    \centering
    \begin{tabular}{|p{\textwidth}|}%
    \hline
        \footnotesize 14\_words, 14\_worte, 14words, abolition\_of\_the\_german\_people, abolition\_of\_the\_occident, abolition\_of\_white\_people, abschaffung\_des\_abendlandes, abschaffung\_des\_deutschen\_volkes, abschaffung\_des\_deutschen\_volks, auslöschen, ausmerzen, ausrotten, ausrottung, bevölkerungsaustausch, bevölkerungsreduzierung, bilderberg, bilderberg-, bilderberg\_group, bilderberg-club, bilderberg-conference, bilderberg-conferences, bilderberger, bilderberger-conference, bilderberger-konferenz, bilderbergern, bilderbergers, bilderberg-group, bilderberg-gruppe, bilderberg-konferenz, bilderberg-konferenzen, bill\_gates, birth\_jihad, birth-jihad, black\_helicopters, bürgerkriegsflüchtlinge, cemetery\_in\_prague, civil\_war\_refugees, coudenhove-kalergi, culture\_war, deckmantel, deckmäntelchen, deep\_state, der\_große\_austausch, drahtzieher, drahtziehern, edmond\_de\_rothschild, elders\_of\_zion, eradicate, eurabia, eurabien, exterminate, extermination, extinction, fadenzieher, false\_flag, false\_flag\_angriff, false\_flag\_angriffe, false\_flag\_attack, false\_flag\_attacks, false\_flag-angriff, false\_flag-angriffe, financial\_elite, financial\_elites, financial\_lobby, financial\_mafia, finanzelite, finanzeliten, finanzlobby, finanzmafia, flüchtlingsinvasion, foreignation, freemason, freimaurer, friedhof\_in\_prag, geburten-dschihad, geburten-jihad, geheimbund, geheimbünde, geheimgesellschaft, geheimpolitik, genocide, genozid, george\_soros, germoney, global\_replacism, globalism, globalismus, globalist, globalisten, globalists, great\_exchange, great\_replacement, große\_austausch, großem\_austausch, großen\_austausch, großer\_austausch, helter\_skelter, illuminaten, illuminati, islamisierung, islamization, jesuit\_order, jesuiten, jesuitenorden, jesuitism, jesuitismus, jesuits, jewish\_republic, judenrepublik, kalergi\_plan,
        kalergi-plan, killuminati, kulturkampf, 
        marionette, marionetten, marionettenregierung, marionettentheater, massenaustausch\_der\_bevölkerung, mastermind, menschenlenkung, menschheit\_reduzieren, migration\_als\_waffe, migration\_as\_a\_weapon, migration\_weapon, migrationswaffe, multiculturalism, multiculturalist,  multikulturalismus, multikulturalist, neue\_weltordnung, neueweltordnung, new\_world\_order, nwo, one\_world\_government, owg, pegida, people\_management, people's\_death, population\_exchange, population\_reduction, population\_replacement, protocols\_of\_the\_elders\_of\_zion, protocols\_of\_zion, protokolle\_der\_weisen\_von\_zion, protokolle\_zions, puppet, puppet\_government, puppet\_master, puppet\_theater, puppets, race\_genocide, racial\_genocide, rassenmischung, reduce\_humankind, refugee\_invasion, relocation, repopulation, repopulation\_machinery, repopulation\_policy, rothschild, rothschilds, satanic\_world\_elite, satanic\_world\_elites, satanische\_weltelite, satanische\_welteliten, secret\_policy, secret\_societies, secret\_society, sir\_john\_retcliffe, soros, string\_puller, string\_pullers, strippenzieher, strippenzieherin, strippenziehern, the\_octopus, tiefe\_staat, tiefen\_staat, tiefer\_staat, umsiedlung, umvolkung, umvolkungsmaschinerie, umvolkungspolitik, usrael, verausländerung, volksaustausch, volkstod, war\_on\_whites, weise\_von\_zion, weisen\_von\_zion, weißer\_genozid, weltherrschaft, weltjudentum, weltordnung, weltregierung, weltverschwörung, white\_extinction, white\_genocide, white\_replacement, world\_conspiracy, world\_control, world\_domination, world\_government, world\_judaism, world\_order, zionist\_congress, zionist\_lodges, zionist\_occupied\_government, zionist\_protocols, zionistisch\_besetzte\_regierung, zionistische\_logen, zionistische\_protokolle, zionistischer\_kongress, zog, zog-\\
        \hline
    \end{tabular}
    \caption{Expert-selected seed dictionary with 215 English and German conspiracy-related keywords based on a German language dictionary by \cite{puschmannRPCLexDictionaryMeasure2022} and additional literature. While the seed dictionary retains German-language keywords, only English-language 4chan posts were collected in our experiment.}
    \label{tab:4chan_seed_dict}
\end{table}

\begin{table*}[htb!]
    \centering
    \footnotesize
    \begin{tabular}[t]{|l||l|l|}
    \hline
    & Start date & End date\\
    \hline
    \hline
    Week 1& 28.02.2011 & 06.03.2011 \\
    \hline
    Week 2& 15.08.2011 & 21.08.2011 \\
    \hline
    Week 3& 12.03.2012 & 18.03.2012 \\
    \hline
    Week 4& 15.10.2012 & 21.10.2012 \\
    \hline
    Week 5& 07.01.2013 & 13.01.2013 \\
    \hline
    Week 6& 04.11.2013 & 10.11.2013 \\
    \hline
    Week 7& 24.02.2014 & 02.03.2014 \\
    \hline
    Week 8& 04.08.2014 & 10.08.2014 \\
    \hline
    Week 9& 19.01.2015 & 25.01.2015 \\
    \hline
    Week 10& 02.11.2015 & 08.11.2015 \\
    \hline
    Week 11& 30.05.2016 & 05.06.2016 \\
    \hline
    Week 12& 12.09.2016 & 18.09.2016 \\
    \hline
    Week 13& 27.03.2017 & 02.04.2017 \\
    \hline
    Week 14& 21.08.2017 & 27.08.2017 \\
    \hline
    Week 15& 02.04.2018 & 08.04.2018 \\
    \hline
    Week 16& 03.12.2018 & 09.12.2018 \\
    \hline
    Week 17& 22.04.2019 & 28.04.2019 \\
    \hline
    Week 18& 16.12.2019 & 22.12.2019 \\
    \hline
    Week 19& 03.02.2020 & 09.02.2020 \\
    \hline
    Week 20& 26.10.2020 & 01.11.2020 \\
    \hline
    Week 21& 31.05.2021 & 06.06.2021 \\
    \hline
    Week 22& 11.10.2021 & 17.10.2021 \\
    \hline
    \end{tabular}
    \caption{22 sample weeks (2 weeks in each year from 2011 to 2021) used to collect 4chan data.}
    \label{Tab:4chan_sample_weeks}
\end{table*}

\newpage
\setlength\bibitemsep{0pt}
\printbibliography

\end{document}